\documentclass[10pt,journal,compsoc]{IEEEtran}
\ifCLASSOPTIONcompsoc
  \usepackage[nocompress]{cite}
\else
  \usepackage{cite}
\fi
\usepackage{hyperref}

\ifCLASSINFOpdf
  \usepackage[pdftex]{graphicx}
  \graphicspath{{img/}}
  
\else
\fi

\usepackage{booktabs}
\usepackage{amsfonts}
\usepackage{enumitem}
\usepackage{multirow}
\usepackage{balance}
\usepackage{xcolor,tabularx,colortbl}

\usepackage{amsmath}
\usepackage{amssymb}
\usepackage{physics}
\usepackage[ruled,linesnumbered]{algorithm2e}

\usepackage{graphicx}
\usepackage{caption}
\usepackage{hyperref}

\usepackage{mdframed}

\SetKwInput{KwInput}{Input}
\SetKwInput{KwOutput}{Output}
\SetKwInput{KwParameter}{Parameter}

\usepackage[capitalize]{cleveref}
\crefname{section}{Sec.}{Secs.}
\Crefname{section}{Section}{Sections}
\Crefname{table}{Table}{Tables}
\crefname{table}{Tab.}{Tabs.}

\DeclareMathOperator*{\argmax}{arg\,max}

\begin{document}
%
\title{Visual Reasoning: from State to Transformation
}
%
%
%
%

\author{Xin~Hong,
        Yanyan~Lan*,
        Liang~Pang,
        Jiafeng~Guo,
        and~Xueqi~Cheng,
\IEEEcompsocitemizethanks{\IEEEcompsocthanksitem * YanyanLan is the corresponding author.\IEEEcompsocthanksitem Xin Hong, Jiafeng Guo, and Xueqi Cheng are with CAS Key Lab of Network Data Science and Technology, Institute of Computing Technology (ICT), Chinese Academy of Sciences (CAS), Beijing, China. Email: \{hongxin19b, guojiafeng, cxq\}@ict.ac.cn.
\IEEEcompsocthanksitem Yanyan Lan is with Institute of AI Industrial Research, Tsinghua University, Beijing, China. Email: lanyanyan@tsinghua.edu.cn.
\IEEEcompsocthanksitem Liang Pang is with Data Intelligence System Research Center, Institute of Computing Technology (ICT), Chinese Academy of Sciences (CAS), Beijing, China. Email: pangliang@ict.ac.cn.}

}

\IEEEtitleabstractindextext{%
\begin{abstract}

Most existing visual reasoning tasks, such as CLEVR in VQA, ignore an important factor, i.e.~transformation. They are solely defined to test how well machines understand concepts and relations within static settings, like one image. Such \textbf{state driven} visual reasoning has limitations in reflecting the ability to infer the dynamics between different states, which has shown to be equally important for human cognition in Piaget's theory. To tackle this problem, we propose a novel \textbf{transformation driven} visual reasoning (TVR) task. Given both the initial and final states, the target becomes to infer the corresponding intermediate transformation. Following this definition, a new synthetic dataset namely TRANCE is first constructed on the basis of CLEVR, including three levels of settings, i.e.~Basic (single-step transformation), Event (multi-step transformation), and View (multi-step transformation with variant views). Next, we build another real dataset called TRANCO based on COIN, to cover the loss of transformation diversity on TRANCE. Inspired by human reasoning, we propose a three-staged reasoning framework called TranNet, including observing, analyzing, and concluding, to test how recent advanced techniques perform on TVR. Experimental results show that the state-of-the-art visual reasoning models perform well on Basic, but are still far from human-level intelligence on Event, View, and TRANCO. We believe the proposed new paradigm will boost the development of machine visual reasoning. More advanced methods and new problems need to be investigated in this direction. The resource of TVR is available at \url{https://hongxin2019.github.io/TVR/}.

\end{abstract}

\begin{IEEEkeywords}
Visual Reasoning, Transformation, Visual Understanding, Deep Learning
\end{IEEEkeywords}
}

\maketitle

\IEEEdisplaynontitleabstractindextext

%

\IEEEraisesectionheading{\section{Introduction}\label{sec:introduction}}

%
%
%
%

\IEEEPARstart{V}isual reasoning goes well beyond object recognition, which is the process of solving problems on the basis of analyzing visual information. Although this task is easy for humans, it is tremendously difficult for vision systems, because it usually requires higher-order cognition and reasoning about the world. Recently, several visual reasoning tasks have been proposed and attract lots of attention in the community of artificial intelligence. For example, CLEVR~\cite{johnson2017clevr}, the most representative visual question answering (VQA) task, defines a question answering paradigm to test whether machines have spatial, relational, and other reasoning abilities for a given image. Visual entailment tasks such as NLVR~\cite{suhr2017corpus, suhr2019corpus} ask models to determine whether a given description is true about states of images. Visual commonsense reasoning tasks, such as VCR~\cite{zellers2019vcr}, further require a rationale explaining the predicting answer.

\begin{figure}[t]
    \begin{center}
        {\includegraphics[width=\columnwidth]{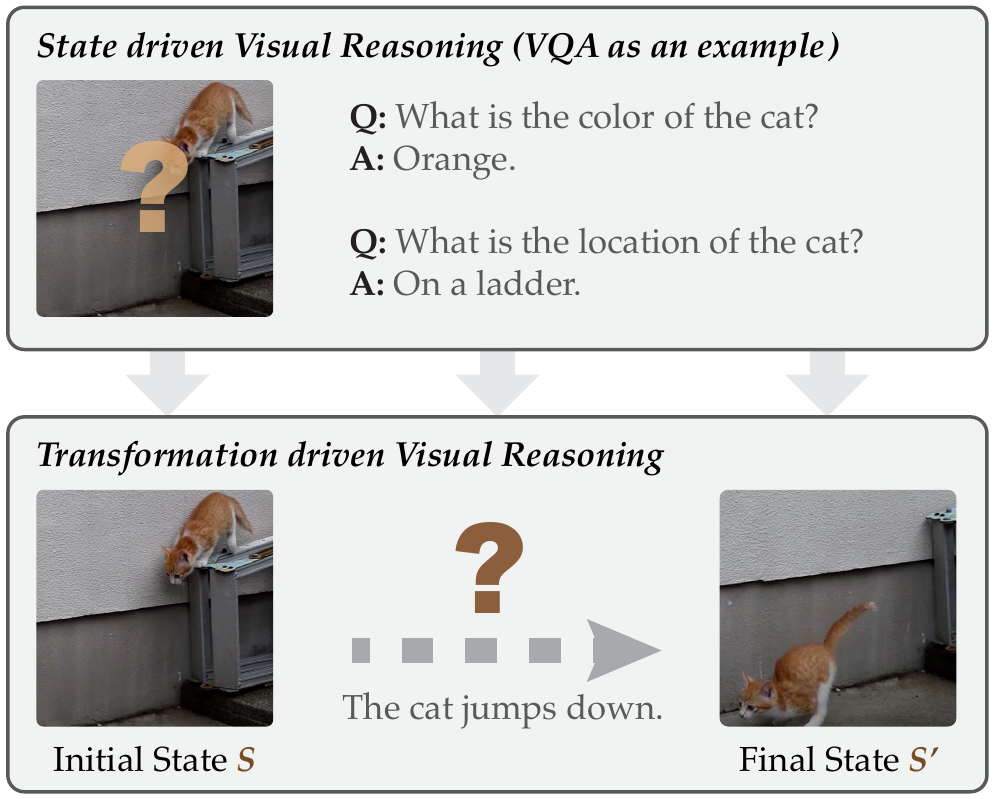}}
    \end{center}
    \caption{State-driven visual reasoning (top) v.s. transformation-driven visual reasoning (bottom).}
    \label{fig:comparison}
\end{figure}

It has been shown from the above description that these visual reasoning tasks are all defined at the {\em state} level. For example, the questions and answers in VQA and VCR as well as the language descriptions in NLVR are just related to the concepts or relations within states, i.e.~an image or images. We argue that this kind of {\em state driven visual reasoning} fails to test the ability to reason dynamics between different states. In the bottom line of \cref{fig:comparison}, the first image shows a cat on a ladder, and in the second image, the same cat is under the ladder. It is natural for a human to reason dynamics here after analyzing, that the cat jumps down the ladder. Piaget's cognitive development theory~\cite{piaget1977role} describes the dynamics between states as transformation, and tells that human intelligence must have functions to represent both the transformational and static aspects of reality. In addition, without modeling transformation, complicated tasks such as visual storytelling~\cite{huang2016visual} and visual commonsense inference~\cite{park2020visualcomet} are hard to be solved, since these tasks involve not only static states but also dynamic transformations, such as actions and events. Though these tasks are closer to reality, they are too complicated to serve as a good testbed for transformation based reasoning. Because these tasks combine too many other requirements, such as recognition and language generation abilities, which makes it hard to independently assess transformation reasoning. Therefore, it is crucial to define a specific task to be able to quantitatively evaluate the ability to reason transformation.

In this paper, we define a novel {\em transformation driven visual reasoning } (TVR) task. Given the initial and final states, like two images, the goal is to infer the corresponding single-step or multi-step transformation. While states are naturally represented as images, the transformation has many choices in its form. Without loss of generality, in this paper, we explore two definitions. In the first definition, transformations are changes of object attributes, therefore a single-step and multi-step transformation are represented as a triplet {\em (object, attribute, value)} and a sequence of triplets, respectively. These triplets, which are basic transformation units, are called \textit{atomic transformations}. In the second definition, atomic transformations are video clips to show the entire change process. Therefore, a single-step and multi-step transformation are respectively represented as a clip of video and a sequence of video clips.

Following the definition of TVR, we first construct a new dataset called TRANCE, to test and analyze how well machines can understand the transformation. We construct TRANCE based on the synthetic dataset CLEVR \cite{johnson2017clevr}, since it is better to first study TVR in a simple setting and then move to more complex real scenarios, just like people first study VQA on CLEVR and then generalize to more complicated settings like GQA. CLEVR has defined five types of attributes, i.e.~color, shape, size, material, and position. Therefore, it is convenient to define the transformation for each attribute, e.g.~the color of an object is changed from red to blue. Given the initial and final states, i.e.~two images, where the final state is obtained by applying a single-step or multi-step transformation on the initial state, a learner is required to well infer such transformation. To facilitate the test for different reasoning levels, we design three settings, i.e.~Basic, Event, and View. Basic is designed for testing single-step transformation. Event and View are designed for more complicated multi-step transformation, where the difference is that View further considers variant views in the final state. \cref{fig:tasks} gives an example of three settings.

The biggest limitation of TRANCE is the small diversity of transformation. Imagine that objects in real life can be transformed into different states through a large number of different transformations. Therefore, we build another dataset called TRANCO to reduce the gap between TRANCE with real, by reasoning transformations on real data. Given such a large transformation space, it is infeasible to list and label all available atomic transformations like TRANCE. As a result, the alternative way is to further require models to generalize to unseen transformations, which is actually the basic requirement for practical applications. TRANCO is thus designed to reason ``open-world''~\cite{Bendale2015TowardsOW} transformations. That is, given the initial and final states, a learner needs to find a sequence of atomic transformations from test candidates, while these test candidates can not be accessed during training. This setting is different from TRANCE since atomic transformations in TRANCE are a constant set of attribute changes on limited objects. In TRANCO, atomic transformations are represented as the aforementioned video clips, so that annotation of existing datasets could be used. Specifically, TRANCO is built on the instructional video dataset COIN, which contains clip annotations that are equivalent to atomic transformations. That is to say, each video contains multiple annotated clips and each clip is corresponding to a step of completing a certain job. The problem then becomes to finding correct video clips given the initial and final frames, while the results are evaluated under the protocol of TVR.

In the experiments, we would like to test how well existing reasoning techniques~\cite{johnson2017inferring, hudson2018compositional} work on TVR. However, since these models are mainly designed for existing reasoning tasks, they cannot be directly applied to TVR. To tackle this problem, we propose a human-inspired reasoning framework specific for TVR, called as TranNet. The design philosophy, as well as the architectural details, are introduced in \cref{sec:model}. In brief, TranNet extracts essential features from two-state images, and then circularly decodes latent representations to predict a sequence of atomic transformations.
With TranNet, existing techniques can be conveniently adapted to TVR. For example, we consider ResNet~\cite{he2016deep}, Bilinear-CNN~\cite{lin2015bilinear}, DUDA~\cite{Park_2019_ICCV}, and CLIP~\cite{Radford2021LearningTV} for encoding, GRU~\cite{cho2014learning}, and Transformer~\cite{vaswani2017attention} for decoding. Experimental results show that deep models perform well on the Basic setting of TRANCE, but are far from human's level on Event, View, and even worse on TRANCO, demonstrating high research potential in this direction.

In summary, the contributions of our work include: 1) the definition of a new visual reasoning paradigm, to learn the dynamics between different states, i.e.~transformation; 2) a new synthetic dataset called TRANCE, to test three levels of transformation reasoning, i.e.~Basic, Event, and View; 3) a real dataset called TRANCO, to test ``open-world'' transformation reasoning; 4) the proposal of a human-inspired transformation reasoning framework TranNet; 5) experimental studies of the existing SOTA reasoning techniques on TRANCE and TRANCO show the challenges of the TVR and some insights for future model designs.

\section{Related Works}

Visual reasoning is an emerging research topic in the field of machine learning, which requires more artificial intelligence than tasks like classification, detection, and captioning. Visual Question Answering (VQA) is the most popular visual reasoning task. Questions in the earliest VQA dataset~\cite{VQA, balanced_binary_vqa, goyal2017making} are usually concerned about the category or attribute of objects. Recent VQA datasets have improved the requirements for image understanding by asking more complex questions, e.g.~Visual7W~\cite{zhu2016visual7w}, CLEVR~\cite{johnson2017clevr}, OK-VQA~\cite{marino2019ok}, and GQA~\cite{Hudson_2019_CVPR}. In addition, two other forms of visual reasoning tasks need to be mentioned. Visual entailment tasks~\cite{suhr2017corpus, suhr2019corpus, xie2018visual, xie2019visual} ask models to determine whether a given description is true about visual inputs. Visual commonsense reasoning~\cite{wang2018fvqa, zellers2019vcr} tasks further require the model to provide a rationale explaining why its answer is right. Solving these tasks is meaningful and requires various reasoning abilities. However, the above tasks are all constrained to be within static states, which ignores the dynamics between different states.

Recently, several new visual reasoning tasks have jointly considered multiple states. For example, CATER~\cite{girdhar2020cater} tests the ability to recognize compositions of object movements, while our task contains more diverse transformations rather than just moving. Furthermore, CATER along with other video reasoning tasks such as CLEVRER~\cite{yi2020clevrer} and physical reasoning~\cite{bakhtin2019phyre, Baradel_2020_ICLR_cophy} is usually based on dense input states, which make the transformations hard to define and evaluate. Before moving to these complex scenarios, our TVR provides a simpler formulation by explicitly defining the transformations between two states, which is more suitable for testing the ability of transformation reasoning. CLEVR-Change~\cite{Park_2019_ICCV}, the most relevant work, requires captioning the change between two images. The novelty is that TVR isolates the ability to reason dynamics from captioning to provide a more thorough evaluation. Furthermore, CLEVR-Change only focuses on single-step transformations.

The concept of transformation has also been mentioned in many other fields. In~\cite{isola2015discovering, nagarajan2018attributes, li2020symmetry}, transformations are used to learn good attribute representations to improve classification accuracy. In~\cite{fathi2013modeling, wang2016actions, liu2017jointly, alayrac2017joint, zhuo2019explainable}, transformations on object or environment are detected to improve the performance of action recognition. However, those works in attribute learning and action recognition fields only consider single-step transformation, thus not appropriate for testing a complete transformation reasoning ability.
Procedure planning~\cite{Chang2020ProcedurePI} has a similar task formulation to ours but we see this problem from different perspectives. TVR motivates transformation as important as the state, while procedure planning specially cares about actions to complete a goal. Specifically, we provide a more comprehensive definition and evaluation for transformation, from synthetic to real, from single-step to multiple-step, and procedure planning can be seen as a special case of TVR.

\section {Task Description} \label{sec:task}


Transformation driven Visual Reasoning (TVR) is a visual reasoning task that aims at testing the ability to reason the dynamics between states. Formally, we denote the state space as $\mathcal{S}$ and the transformation space as $\mathcal{T}$. The transformational process can be illustrated as a function $f:\mathcal{S}\times\mathcal{T}\to\mathcal{S}$, which means a state is transformed into another state under the effect of a transformation. And our task is defined as:

\mdfdefinestyle{TaskFrame}{
    linecolor=gray,
    linewidth=0.25pt,
    innertopmargin=3pt,
    innerbottommargin=3pt,
    innerrightmargin=3pt,
    innerleftmargin=3pt,
    leftmargin = -2pt,
    rightmargin = -2pt,
    backgroundcolor= black!03}

\begin{mdframed}[style=TaskFrame]

\textbf{Transformation Driven Visual Reasoning:}

\noindent$\mathcal{S}$ is the state space, and $\mathcal{T}$ is the transformation space.

\noindent\textbf{Input:} 
\begin{itemize}
    \item the initial state $S\in\mathcal{S}$, represented as an image,
    \item the final state $S' \in\mathcal{S}$, represented as an image.
\end{itemize}

\noindent\textbf{Output:} A transformation $T \in \mathcal{T}$, so that $f (S, T) = S'$.


\end{mdframed}


With this definition, most existing state driven visual reasoning tasks can be extended to the corresponding transformation driven ones. For example, the VQA task, such as CLEVR, can be extended to ask about the transformation between two given images, with answers as the required transformation. In the extension of NLVR, the task becomes to determine whether a sentence describing the transformation is true about the two images, e.g.~the color of the bus is changed to red. Since TVR itself is defined as an interpretation task, we do not need any further rational explanations, and the extension of VCR will stay the same as CLEVR. We can see that the intrinsic reasoning target of these tasks is the same, that is to infer the correct transformation, while the difference lies in the manifestation.


In TVR, states are naturally represented as images to capture static moments, but the transformation has many choices in its form. For example, any changes in pixel value can be treated as a transformation, but this representation is meaningless for humans. Another way to describe transformation is natural language~\cite{Park_2019_ICCV}. However, natural language is not precise and sometimes ambiguous, making it difficult to evaluate the accuracy of the predicted transformations.

In this paper, we explore two transformation definitions. In the first definition, transformations only affect limited attributes with limited options just like~\cite{Park_2019_ICCV}, but the form is changed from the caption to a more concrete one, i.e.~attribute-level change of an object, represented as a triplet $(o, a, v)$, which means the object $o$ with the attribute $a$ is changed to the value $v$. Except for the representation, another limitation of~\cite{Park_2019_ICCV} is they only consider single attribute changes between states, while multiple attribute changes could exist between states in practice. A more general formulation should consider multiple transformations as well as their order. To be clear, a basic transformation such as the triplet $(o, a, v)$ is called an \textit{atomic transformation}, denotes as $t$. And the transformation $T$, denotes as a sequence of atomic transformations that $T=\{t_1, t_2, \dots, t_n\}, t_i =(o_i, a_i, v_i)\in \mathcal{T_A}$, where $n$ is the number of atomic transformations, and $\mathcal{T_A} \subset \mathcal{T}$ is the atomic transformation space.

In more complex scenarios, such as in real data,
one single atomic transformation may affect multiple attributes. Take the cat example again, a simple jumping affects at least the location and the pose of the cat. It is not suitable to represent transformations as attribute changes in this situation.
Instead, representing an atomic transformation as a clip of video, completely showing the whole change process is natural and more friendly for annotating. The definition of the transformation keeps the same as $T=\{t_1, t_2, \dots, t_n\}$, while $ t_i=c_i \in \mathcal{T_A}$ and $c_i$ is a clip from a video.

Different definitions of transformation can lead to different ways of evaluation. The most ideal way of evaluating the prediction $\hat{T}$, is to first obtain the corresponding simulated final state $\hat{S}'=f(S, \hat{T})$, and then check whether $\hat{S}'$ is the same as ground truth final state $\hat{S}$. The first definition that represents transformations as attribute changes of objects is appropriate for this evaluation. However, in real scenarios, it is hard to obtain a simulated final state. We have defined the transformation as a sequence of clips. The goodness of this definition is annotating-friendly, but it is limited for the real data that the evaluation could only be done by comparing predicting $\hat{T}$ with the given reference transformation $T$. The problem here is that $T$ may not be the only way in practice to transform the state from $S$ into $S'$, thus the evaluation is imperfect. \cref{sec:trance_metrics} and \cref{sec:tranco_metrics} will introduce the detailed evaluation protocols for TRANCE and TRANCO.

\section{Synthetic Data: TRANCE} \label{sec:dataset}

We first study TVR under the synthetic setting, in which we build a new data set by extending CLEVR, namely TRANCE (\underline{Tran}sformation on \underline{C}L\underline{E}VR). Besides, we describe how to define proper TVR objectives and corresponding evaluation protocols with respect to TRANCE.

\subsection{Dataset Setups}


CLEVR~\cite{johnson2017clevr} is a popular VQA dataset, which first introduces the concept of visual reasoning. The target of CLEVR is to answer questions about counting, comparing, logical reasoning, and so on, according to given images. The content of images is about simple objects, such as cubes, spheres, and cylinders, which have different sizes, materials, and colors. Specifically, for each object, there are 3 shapes, 2 sizes, 2 materials, 8 colors, and infinity locations to be selected, as listed in \cref{tab:values} annotated with *.

With so many attributes that are convenient to be modified, we can easily define atomic transformations as changes of these attributes on objects. This is the major reason that we choose CLEVR to extend. Another reason is that images can be synthesized using Blender~\cite{blender2016blender} with small costs. Therefore, it is practicable to create millions of samples.

\begin{table}[t]
    \centering
    \caption{Attributes and values in TRANCE.}
    \centerline{\includegraphics[width=0.95\columnwidth]{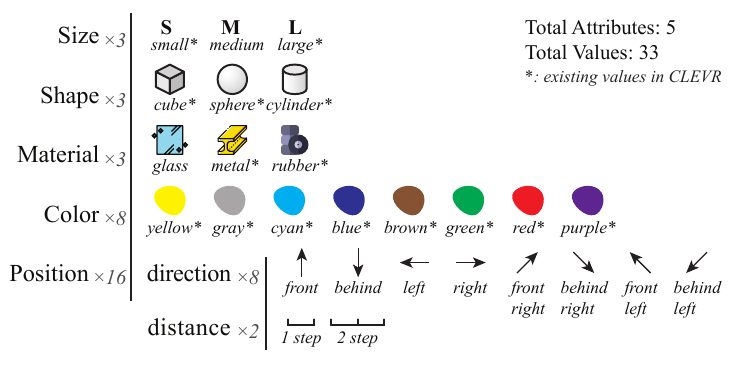}}
    \label{tab:values}
\end{table}

CLEVR provides a good foundation on attributes and values, which are fundamental items of the atomic transformation triplet $(o, a, v)$, as we introduced in \cref{sec:task}. However, the distance to defining atomic transformations well still exists unless we proceed with several modifications or designs. The first problem is how to represent an object in the answer. Existing methods such as CLEVR and CLEVR-Change use text which has ambiguity issues making the evaluation unreliable, while CLEVR-Ref+~\cite{liu2019clevr} employs bounding boxes that are specific but require the additional ability of detection. Therefore, we design to provide additional information, which is the attributes of the initial objects, including the index, color, material, and other attribute values. In this way, an object can be referred to with its index. Note machines still need to perform their own recognition to align objects in images with given attributes. The second problem is available values in size and material are too few, therefore we add medium size and glass material. The last problem is the available values of position transformation are infinite in the space of $\mathbb{R}^2$, which is not computational friendly. To reduce the available values, we change the position from absolute values into relative values by using direction and step to represent the position transformation. Specifically, we consider eight directions as shown in~\cref{tab:values}. In addition, we define a coordinate system, in which $x$ and $y$ are both restricted to $[-40, 40]$, and objects can only be placed on integer coordinates. The moving step can be valued as 1 or 2, where 1 step equals 10 in our coordinate system. Except for normal moving action, we are also interested in whether the vision system could understand actions like moving in and moving out, so the plane is split, where the visible area is at the center and the invisible area is around the visible area, and the moving in and out operations can be defined correspondingly. To be reasonable, objects shouldn't be overlapped and moved out of the plane during transformation.

Having defined the atomic transformation, we will now move on to introduce how to generate samples. The first step is the same as CLEVR, which is randomly sampling a scene graph. According to the scene graph, CLEVR then generates questions and answers with a functional program and renders the image with Blender. Different from CLEVR, the next step in TRANCE becomes randomly sampling a sequence of atomic transformations, where the length ranges from 1 to 4, which is called the \textit{reference transformation}. By applying the reference transformation to the initial scene graph, we obtain the final scene graph. At last, two scene graphs are rendered into images ($h:240\times w:320$).

To reduce the potential bias from random sampling, we carefully control the sampling process of scene graph and transformation by balancing several factors. In scene graph sampling, we balance objects' attributes and the number of visual objects in the initial state. In transformation sampling, the length of the transformation, the object number, n-gram atomic transformation, and the move type are all balanced. Throughout all elements, N-gram atomic transformation is the hardest to be balanced and it refers to the sub-sequence of atomic transformations with the length of $n$. By balancing these factors, we reduce the possibility that a learner utilizes statistics features in the data to predict answers. In the supplementary material, we show the statistics of the dataset and our balancing method in detail.


\begin{figure}[t]
    \begin{center}
        {\includegraphics[width=\columnwidth]{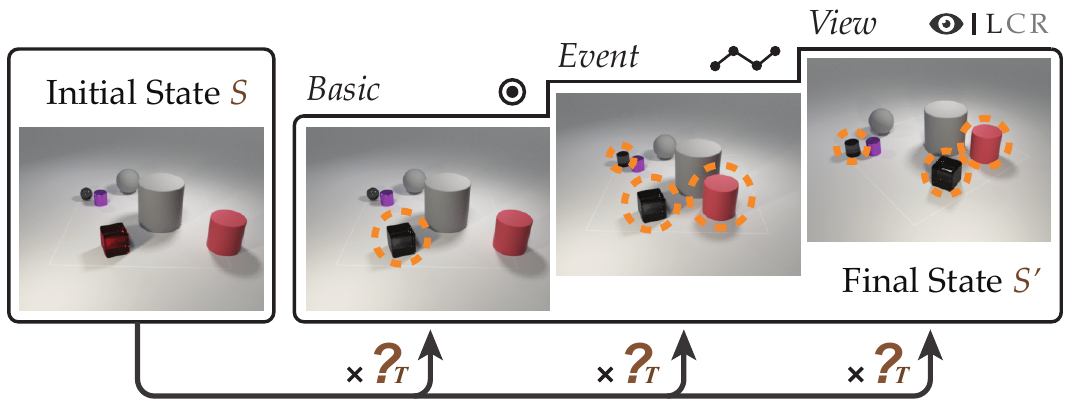}}
    \end{center}
    \caption{Illustration of three settings in TRANCE. \textbf{Basic:} Find the single-step transformation between the initial and final state. \textbf{Event:} Find the multi-step transformation between two states. \textbf{View:} Like Event, but the view of the final state is randomly selected from \textit{Left}, \textit{Center}, and \textit{Right}.}
    \label{fig:tasks}
\end{figure}

\subsection{Three Levels of Settings} \label{sec:settings}
We design three settings, i.e.~Basic, Event, and View, to facilitate the study on different levels of transformation reasoning. Basic is first designed for single-step transformation and then Event is for multi-step transformation. To further evaluate the ability of reasoning transformation under a more real condition, we extend Event with variant views to propose View. \cref{fig:tasks} shows three different settings and more examples can be found in the supplementary material.

\textbf{Basic.}
Basic is a simple problem designed to mainly test how well a learner understands atomic transformations. The target of Basic is to infer the single-step transformation between the initial and final states. That is, given a pair of images, the task is to find out which attribute $a$ of which object $o$ is changed to which value $v$. We can see that this task is similar to the previous game ``Spot the Difference"~\cite{jin2013spoid}, in which the player is asked to point out the differences between two images. However, Basic is substantially different from the game. Basic cares about the object level differences while the game focuses on the pixel level differences. Therefore, Basic can be viewed as a more advanced visual reasoning task than the game.

\textbf{Event.}
Considering only the single-step transformation is obviously not enough. In reality, it is very common that multi-step transformation exists between two states. Therefore, we construct this multi-step transformation setting to test whether machines can handle this situation. The number of transformations between the two states is randomly set from 1 to 4. The goal is to predict a sequence of atomic transformations that could reproduce the same final state from the initial state. To resolve this problem, a learner must find all atomic transformations and arrange them correctly. Compared with Basic, it is possible to have multiple atomic transformations, which improves the difficulty of finding them all. Meanwhile, the order is essential in the Event because atomic transformations may be dependent. For example, moving A first and then moving B to A's place is non-exchangeable, otherwise, B will overlap A.

\textbf{View.}
In real applications, the angle of observation is not fixed like in Basic and Event. To tackle this problem, we extend the Event setting to View, by capturing two states with cameras in different positions. In practice, for simplicity but without loss of generality, we set three cameras, placed on the left, center, and right sides of the plane. The initial state is always captured by the center camera, while for the final state, images are captured with all three cameras. Thus, for each sample, we obtain three pairs for training, validation, and testing with the same initial state but different views of the final states. With this design, it is possible to evaluate how well a vision system understands object-level transformation under variant views.

\subsection{The Evaluation Protocol}
\label{sec:trance_metrics}

For the single-step transformation setting, i.e.~Basic, the answer is unique. Therefore, we can evaluate the performance by directly comparing the prediction with the reference transformation. Specifically, in the TRANCE dataset, we consider fine-grained accuracy and overall accuracy.

{\em ObjAcc, AttrAcc, ValAcc}. Fine-grained accuracy corresponds to three  elements in the  transformation triplet, including object accuracy (ObjAcc),  attribute accuracy (AttrAcc), and value accuracy (ValAcc).

{\em Acc}. The overall accuracy (Acc) only counts the absolutely correct transformation triplets.

For multi-step transformation settings, i.e~Event and View, it is not suitable to use the above evaluation metrics, since there may exist multiple feasible answers. This is because exchanging some steps like color transformation and shape transformation is acceptable and the final state keeps unchanged. Benefiting from the simple setting of TRANCE, it is convenient to evaluate the predicted transformation by simulation. Specifically, we input the item of predicted transformation sequence $\hat{T} = \{\hat{t_1}, \hat{t_2}, \cdots, \hat{t_n}\}$ one by one to transform the initial state to the simulated final state $\hat{S}'$, i.e.~$S \times \hat{T} \rightarrow \hat{S'}$. A {\em distance} can be computed by counting the attribute level difference between two final states, i.e.~$\hat{S'}$ and $S'$. If the intermediate states do not violate the pre-defined two constraints, including no overlapping and no moving out of the plane, and the distance is zero, then the sequence is \textit{correct}. If we ignore the two constraints, which means the order of the sequence is ignored, and find the distance is zero, then the sequence is called \textit{loose correct}.

{\em AD, AND}. A {\em normalized distance} is a distance that is normalized by the length of the reference transformation. {\em AD} and {\em AND} are the average distance and average normalized distance over all samples, respectively.

{\em Acc, LAcc}. The accuracy is the proportion of \textit{correct} samples, while the loose accuracy is the proportion of \textit{loose correct} samples without considering the order:
\begin{equation}
\begin{aligned}
& Acc = \sum_i^m \frac{1}{m}[T_i \ \text{is \textit{correct}}], \\ & LAcc = \sum_i^m \frac{1}{m}[T_i \ \text{is \textit{loose correct}}],
\end{aligned}
\end{equation}
where $m$ is the total number of test samples.

{\em EO}. At last, to measure how well the right order is assigned when all atomic transformations have been found, the error of order $EO = (LAcc - Acc) / LAcc$ is computed.



\begin{figure}[t]
    \centering
    \centerline{\includegraphics[width=0.95\linewidth]{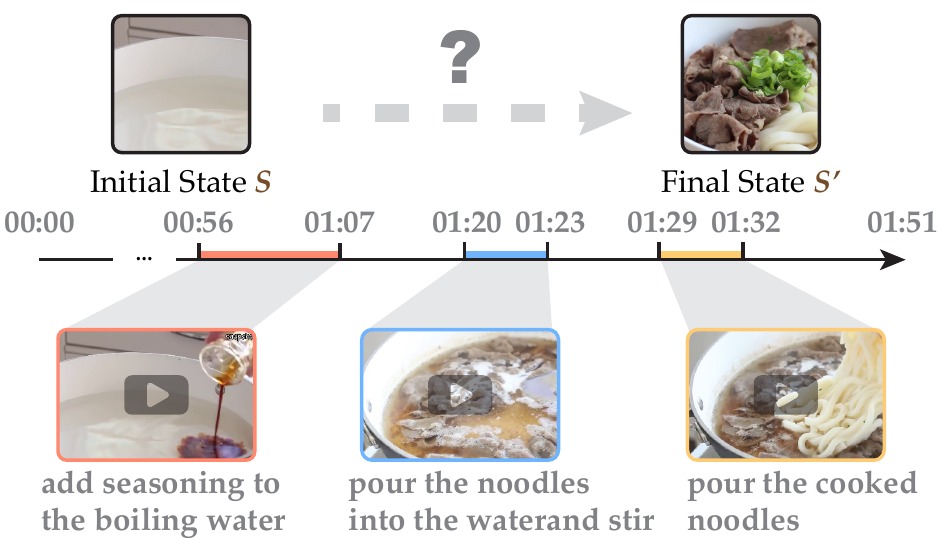}}
    \caption{Illustration of an example from TRANCO. The target is to find a sequence of video clips between the initial and the final state.}
    \label{fig:coin}
\end{figure}

\section{Real Data: TRANCO}

In addition to the synthetic data, we build a real dataset called TRANCO (\underline{Tran}sformation on \underline{CO}IN), to explore the potential role of visual reasoning research in real scenarios.

\subsection{Data Setups}
TRANCO is built based on a well-known comprehensive instruction video data, namely COIN~\cite{Tang2019COINAL}, which consists of 11,827 YouTube videos covering 180 different tasks in daily activities. COIN is widely used in instructional video analysis tasks, including step localization, action segmentation, procedure localization, task recognition, and step recognition. Each video of COIN is comprised of a series of steps annotated with temporal boundaries and descriptions. For example, \cref{fig:coin} shows three main steps of cooking noodles, where each step is represented as a video clip along with a sentence to describe the step.

We choose COIN to build our real dataset for two major reasons. Firstly, videos in COIN are real data covering various daily activities, which meets our requirement of diverse transformations. Furthermore, the step annotations can be reused to reduce the cost of building the dataset, since the steps in COIN are equivalent to our atomic transformations.

As we discussed in \cref{sec:task}, the second transformation definition represents an atomic transformation as a video clip,  which is more suitable for real complex scenarios than attribute-level changes. Under this definition, for each sample in COIN, step video clips are directly transferred to be atomic transformations. The additional elements that we need to construct are the initial and the final states. In practice, the state before these steps is the initial state, and the state after is the final state. Therefore, the first frame of the first step video clip becomes the initial state and the last frame of the last step video clip becomes the final state.

In addition, to simplify the problem, videos containing more than 7 steps are not used, resulting in 11,105 videos and 38778 total video clips. These videos are separated into 8651 train samples, 1024 validation samples, and 1430 test samples. The detailed video distribution on the clip number is shown in \cref{tab:stat_tranco}.

\begin{table}[t]
\setlength{\tabcolsep}{0.35em}
\begin{center}
\begin{small}
\caption{Statistics of TRANCO.}
\label{tab:stat_tranco}
\begin{tabular}{lrrrrrrrrr}
\toprule
& \multirow{2}{*}{videos} & \multirow{2}{*}{clips} & \multicolumn{6}{c}{videos with $k$ clips} \\
\cmidrule{4-9}
&&& $k=$2 & $k=$3 & $k=$4 & $k=$5 & $k=$6 & $k=$7 \\
\midrule
Train & 8651 & 30244 & 2451 & 2497 & 1874 & 947 & 554 & 328 \\
Val   & 1024 & 3616 & 283 & 291 & 235 & 96 & 76 & 43 \\
Test  & 1430 & 4918 & 432 & 425 & 279 & 154 & 87 & 53 \\
\midrule
Total & 11105 & 38778 & 3166 & 3213 & 2388 & 1197 & 717 & 424    \\
\bottomrule
\end{tabular}

\end{small}
\end{center}
\end{table}


\subsection{The Problem Setting}

The goal of TRANCO is to reason ``open-world''~\cite{Bendale2015TowardsOW} transformations, that is, models should generalize to unseen transformations. Specifically, for each video from COIN, two images are given as the initial and the final state respectively, and the target is to find out the original sequence of video clips between the two states as the transformation, from a candidate set of video clips. During testing, the candidate video clips are comprised of all video clips from the testing set and are not exposed to training. With this design, we expect models to adapt to the diverse characteristics of transformations in the real world.


TRANCO is intuitively more difficult than TRANCE. The major difficulty is the objective, i.e.~reasoning ``open-world'' transformation, which requires additional ability to transfer into unseen atomic transformations. Another difficulty comes from the requirement of the higher recognition ability to represent real images or videos. Experiments in \cref{sec:exp_tranco} also confirm these two major difficulties of TRANCO.


\subsection{The Evaluation Protocol}
\label{sec:tranco_metrics}

As we discussed in \cref{sec:task}, the definition of transformation can affect the way of evaluation. In order to determine whether the predicted transformation is correct, it is not feasible to compare simulated final state $\hat{S'}$ with the ground truth final state $S'$ here, since it is hard to simulate the real transformation in TRANCO. The alternative way is to directly compare the predicted transformation $\hat{T}$ with the reference transformation $T$. Nevertheless, it is acceptable for TRANCO, since the steps in instructional videos are usually unique and can not be exchanged. We consider four metrics for evaluation, including the overall exact match rate, two metrics on the ability to find correct atomic transformations without considering the order, and one especially for order assessment. These metrics are introduced in the following.




\textit{Exact Match Rate (EMR).} The first metric is exact match rate, which evaluates the overall performance. It reflects how many predicted transformations are exactly the same as reference transformations, which requires not only the atomic transformations but also the order are exactly the same. We use the exact match rate here to distinguish with the \textit{Acc} in TRANCE, since the meaning and the evaluation method are different.

\textit{Recall, Precision.} These two metrics both concern the ability to find correct atomic transformations and ignore the order of predicted transformations. Recall reflects how many atomic transformations in the reference transformation are found, while precision reflects how many predicting atomic transformations are right. They are given by:
\begin{equation}
    \text{Recall} = \frac{|T \cap \hat{T}|}{|T|}, \quad \text{Precision} = \frac{|T \cap \hat{T}|}{|\hat{T}|}.
\end{equation}


\textit{KTD.} In contrast to recall and precision, KTD (Kendall's-$\tau$ distance) only focuses on order evaluation to reflect how well models sort atomic transformations. KTD is a commonly used metric in the field of information retrieval to evaluate ranking models, the detail can be found in~\cite{kendall1938}. When computing KTD, we only consider the order of intersected atomic transformations $T \cap \hat{T}$. We define that $\text{KTD} = 1$ if $T \cap \hat{T} = \emptyset$.


\textit{SD, NSD.} Similar to TRANCE, we provide step difference and normalized step difference to reflect how well models estimate the number of steps between the initial and final states. SD is the absolute difference between the number of predicted steps and the number of ground truth steps. NSD is the normalized SD, which is the ratio of SD to the number of ground truth steps.

\begin{figure}[t]
    \centering
    \centerline{\includegraphics[width=0.95\linewidth]{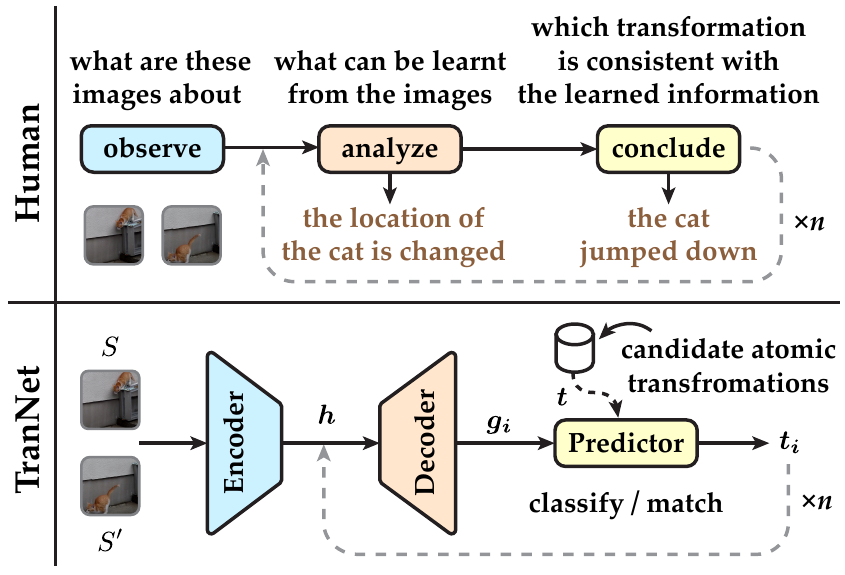}}
    \caption{What we understand about human transformation reasoning (top).  Inspired TranNet framework (bottom).}
    \label{fig:inspiration}
\end{figure}

\section{The TranNet Framework}
\label{sec:model}

In this section, we propose a general framework to tackle the transformation driven visual reasoning problem, including both synthetic and real scenarios.



\subsection{The Basic Idea}

TranNet is inspired by the OODA decision loop theory~\cite{osinga2007science} and our study about how human reason transformation reasoning during human experiments. In~\cref{fig:inspiration}, the top row shows the three stages that we understand about the reasoning process. To reason the transformation from states, a human will first observe images, and then circularly analyze image contents and conclude transformations. Take the cat jumping as an example, a human will first observe to know that these two images are about a cat. Next, the one analyzes the two images and finds out the location of the cat is changed. At last, the one searches mind for a feasible transformation that could explain the finding state change, which is ``the cat jumps down". If the transformation process is complex, e.g. cooking noodles, that a single-step transformation is not enough to complete the entire state changes, one will repeat analyzing and concluding until working out a sequence of transformations as the explanation.

We transform the three stages accordingly into modules as shown in the bottom row of \cref{fig:inspiration}, including encoder, decoder, and predictor, to form the TranNet framework. In the following, we briefly introduce how these modules work and show the instantiations of TranNet on our two problems in \cref{sec:model_tranco} and \cref{sec:model_trance}.

\textbf{Encoder.}
The goal of the encoder is to extract effective features from image pairs, which are mainly associated with the content within and the relation between two states. 
Specifically, an encoder $E$ extracts image features $\vb*{h}$ from two states images $S$ and $S'$:
\begin{equation}
\vb*{h} = E(S, S').
\end{equation}
As for the pair inputs, there are two common ways to extract features from them, i.e.~early fusion and latter fusion. In the early fusion way, input images interact before sending into the network, while in the latter fusion, images are first separately encoded and then interacted at the feature level. The backbone of the encoder can be any common image encoder, such as ResNet~\cite{he2016deep} and Vision Transformer~\cite{dosovitskiy2021an}.

\textbf{Decoder.}
The decoder is a bridge between the encoder and the predictor. The goal of the decoder is to circularly decode information from image representation for the predictor to predict atomic transformations. In the $i$th step, in addition to $\vb*{h}$ from the encoder, the decoder also accepts previous atomic transformations  $\vb*{t_{k<i}}$ as inputs:
\begin{equation}
\label{eq:decoder}
\vb*{g_i} = D(\vb*{h}, \vb*{t_{0}}, \cdots, \vb*{t_{i-1}}),
\end{equation}
where $\vb*{t_{0}}$ is the initial atomic transformation, which could be set by different strategies, e.g. a random initialized vector optimized during learning. RNNs (e.g.~GRU~\cite{cho2014learning}) and transformer~\cite{vaswani2017attention} are selected as two variants of decoders, which are commonly used techniques for sequence generation.

\textbf{Predictor.}
The predictor is responsible for translating the information from the decoder into one specific atomic transformation, which should belong to the candidate atomic transformations. This is implemented by finding $\vb*{t}\in \mathcal{T_A}$ that maximizes the \textit{score} for received $\vb*{g_i}$:
\begin{equation}
\vb*{t_i} = \argmax_{\vb*{t}\in \mathcal{T_A}} score(\vb*{g_i}, \vb*{t}).
\end{equation}
In general, there are two ways to implement the score function, corresponding to two different problem formulations. The first way regards the score as a \textit{classification} function, which maximizes the likelihood of desired atomic transformation given $\vb*{g_i}$. The second one is a \textit{contrastive learning} way, which is to maximize the similarity between the $\vb*{g_i}$ and $\vb*{t}$. The major difference is that the labels or candidates in the classification problem must be fixed and shared between training and testing while contrastive learning does not require this. Therefore, the first way is more suitable for problems with few labels and the second way has more advantages in its generalization ability. The second difference is that the contrastive way needs an extra encoder to encode $\vb*{t}$ so that the similarity between $\vb*{g_i}$ and $\vb*{t}$ can be computed in the same vector space.

Having introduced the basic idea of TranNet, the following two sections discuss how to implement TranNet in two specific scenarios, i.e. TRANCE and TRANCO.

\begin{figure}[t]
    \centering
    \centerline{\includegraphics[width=0.95\linewidth]{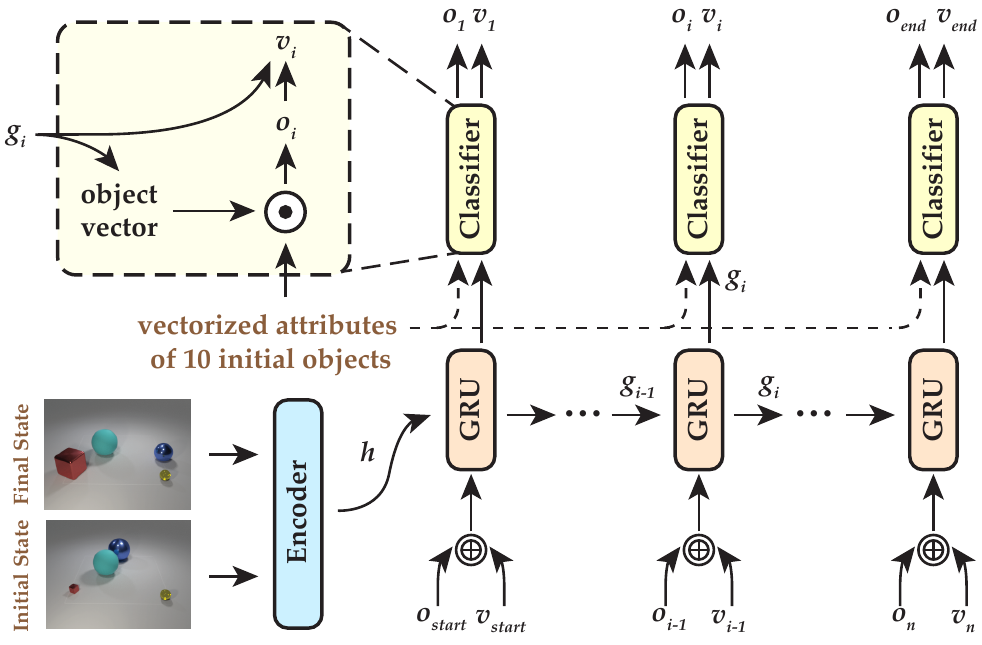}}
    \caption{The architecture of TranceNet.}
    \label{fig:trance_network}
\end{figure}

\subsection{TranceNet}  \label{sec:model_trance}

There are two guidelines we follow to design TranceNet for TRANCE. The first one is to design effective encoders, therefore we compare encoders with different encoding ways and architectures. The second one is to formulate the prediction as a classification problem since the atomic transformation space in TRANCE is fixed.

\cref{fig:trance_network} shows the architecture of TranceNet. In the encoder part, we consider two types of early fusion encoders and two types of latter fusion encoders. Early fusion ways include subtracting ($-$) or concatenating ($\oplus$) two images before feeding them into the networks, such as vanilla CNN or ResNet. We use the network name with a subscript to denote early fusion encoders, for example, ResNet$_-$ means ResNet feeding in subtracted image pairs. The latter fusion encoders include BCNN~\cite{lin2015bilinear} and DUDA~\cite{Park_2019_ICCV}. BCNN is a classical model for fine-grained image classification to distinguish categories with small visual differences. DUDA is originally proposed for change detection and captioning. The main difference between BCNN and DUDA lies in the way of feature-level interaction. We choose GRU~\cite{cho2014learning} and transformer as two different decoders for comparison. The GRU unit updates the hidden state and receives only the last step of atomic transformation and \cref{eq:decoder} becomes:

\begin{equation}
\vb*{g_i} = D(\vb*{g_{i-1}, t_{i-1}}),
\end{equation}
where $\vb*{g_{0}} = \vb*{h}$, and $\vb*{t_{0}}$ is a learned variable. Since the atomic transformations space of TRANCE is fixed as ten objects times all attribute values, it is better to formulate the problem in a classification way, and the final loss function is simplified as a combination of two cross-entropy losses for object and value respectively, represented as:
\begin{equation}
\begin{split}
\mathcal{L} = -\frac{1}{n} \sum_{i=1}^{n} (\vb*{t_{i}^{o}} \cdot \log\vb*{{g}_{i}^{o}} + \vb*{t_{i}^{v}} \cdot \log\vb*{{g}_{i}^{v}}).
\label{eq:loss_trance}
\end{split}
\end{equation}
Note the attribute in the triplet is implied by the value, since each value only belongs to one specific attribute here.

\begin{figure}[t]
    \centering
    \centerline{\includegraphics[width=0.8\linewidth]{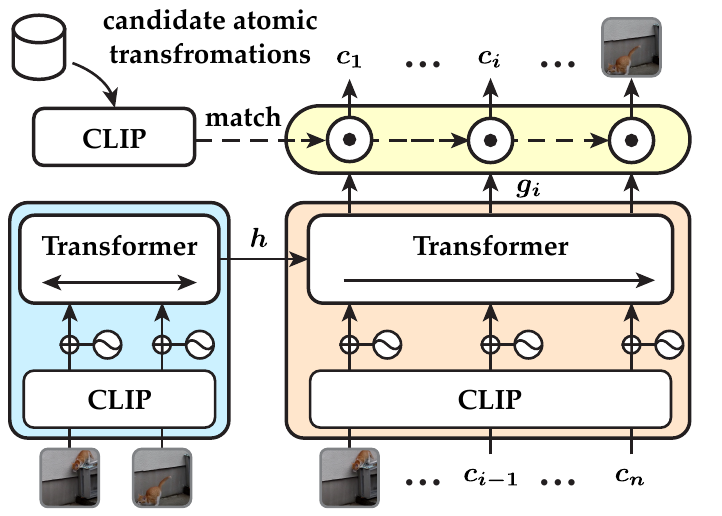}}
    \caption{The architecture of TrancoNet.}
    \label{fig:tranco_network}
\end{figure}

\begin{table*}[t]
\centering
\caption{Model and human performance on Basic, Event, and View. $\Delta$Acc is the accuracy difference between View and Event.}
\setlength{\tabcolsep}{0.45em}
\label{tab:results}
\begin{small}
\begin{tabular}{lrrrrrrrrrrrrr}
\toprule
\multirow{2}{*}{Model} & \multicolumn{4}{c}{Basic} & \multicolumn{4}{c}{Event} & \multicolumn{4}{c}{View} & \multirow{2}{*}{$\Delta$Acc$\uparrow$} \\
 \cmidrule[.5pt](lr){2-5}  \cmidrule[.5pt](lr){6-9}  \cmidrule[.5pt](lr){10-13}
 & ObjAcc$\uparrow$ & AttrAcc$\uparrow$ & ValAcc$\uparrow$ & Acc$\uparrow$ & AD$\downarrow$ & AND$\downarrow$ & LAcc$\uparrow$ & Acc$\uparrow$ & AD$\downarrow$ & AND$\downarrow$ & LAcc$\uparrow$ & Acc$\uparrow$ &  \\
\midrule
CNN$_{-}$-G & 0.9596 & 0.9954 & 0.9834 & 0.9440 & 1.5842 & 0.5217 & 0.4568 & 0.4419 & 2.2649 & 0.8851 & 0.2376 & 0.2300 & -0.2119 \\
CNN$_{\oplus}$-G & 0.9570 & 0.9942 & 0.9798 & 0.9390 & 1.4416 & 0.4725 & 0.4961 & 0.4797 & 2.0671 & 0.7887 & 0.2889 & 0.2789 & -0.2008 \\
BCNN-G & 0.9684 & 0.9946 & 0.9818 & 0.9524 & 1.1299 & 0.3623 & 0.5847 & 0.5610 & 1.2915 & 0.4437 & 0.4977 & 0.4749 & -0.0861 \\
DUDA-G & 0.9534 & 0.9922 & 0.9838 & 0.9394 & 1.3184 & 0.4170 & 0.5612 & 0.5401 & 1.4943 & 0.5130 & 0.4837 & 0.4645 & -0.0756 \\
ResNet$_{-}$-G & 0.9808 & \textbf{0.9982} & 0.9934 & 0.9744 & 1.0072 & 0.3108 & 0.6350 & 0.6057 & 1.0552 & 0.3564 & 0.5704 & 0.5454 & -0.0603 \\
ResNet$_{\oplus}$-G & \textbf{0.9856} & 0.9980 & \textbf{0.9954} & \textbf{0.9814} & 1.0624 & 0.3336 & 0.6217 & 0.5932 & 1.1353 & 0.3760 & 0.5681 & 0.5426 & -0.0507 \\
ResNet$_{-}$-T & - & - & - & - & \textbf{0.8389} & \textbf{0.2601} & \textbf{0.6865} & \textbf{0.6553} & \textbf{0.8832} & \textbf{0.2933} & \textbf{0.6324} & \textbf{0.6012} & -0.0541 \\
ResNet$_{\oplus}$-T & - & - & - & - & 0.8873 & 0.2777 & 0.6743 & 0.6424 & 0.9260 & 0.3084 & 0.6243 & 0.5927 & \textbf{-0.0497} \\
\midrule
Human & 1.0000 & 1.0000 & 1.0000 & 1.0000 & 0.3700 & 0.1200 & 0.8300 & 0.8300 & 0.3200 & 0.0986 & 0.8433 & 0.8433 & 0.0133 \\
\bottomrule
\end{tabular}
\end{small}
\end{table*}

\subsection{TrancoNet} \label{sec:model_tranco}

The requirement to the TrancoNet is higher than TranceNet. Compared with TranceNet, our first guideline additionally requires high recognition ability, therefore we employ pretrained CLIP. The second guideline is to formulate the transformation prediction in a contrastive learning style, because the atomic transformation space of TRANCO are dynamic rather than fixed from training to testing.

\cref{fig:tranco_network} show the architecture of TrancoNet. We choose transformer as the main backbone to better model the order of atomic transformations. Meanwhile, we use a pretrained CLIP image encoder to reduce the training cost of extracting features from the real image and video data. CLIP is pretrained on massive image-text pairs and achieves SOTA on many multi-modal tasks, including video retrieval~\cite{PortilloQuintero2021ASF}. In the encoder part, we only consider the latter fusion way, since early fusion changes the input space and it is impossible to obtain a good performance without tuning CLIP models. The input images are first separately encoded with CLIP image encoder, and then interacted with a transformer encoder. In the decoder part, in $\vb*{i}$th step, a transformer predicts the latent representation $\vb*{g_i}$ by applying cross attention to the state representation $\vb*{h}$ and previous steps of atomic transformations $\{\vb*{c_0} \cdots \vb*{c_{i-1}}\}$, where $\vb*{c_0}$ is chosen to be the initial state and $\vb*{c_{i-1}}$ is the $\vb*{i-1}$th video clip. Refer to~\cite{PortilloQuintero2021ASF}, video clips are also encoded with CLIP image encoder by averaging the encoding results of sampled video frames. Finally, in the predictor part, since the task is more like a problem of ranking the candidates and the candidates are different between training and testing, it is more natural to formulate the problem in a contrastive learning way that maximizes the similarity between $\vb*{g_i}$ and corresponding encoded reference clip $\vb*{c_i}$, while decreasing the similarity with other video clips from candidates:
\begin{equation}
    \mathcal{L} = - \frac{1}{n} \sum^{n}_{i=1} \log \frac{\exp (\vb*{g_i} \cdot \text{CLIP}(\vb*{c_i}) / \tau)}{\sum_{\vb*{c} \in \mathcal{T_A}} \exp(\vb*{g_i} \cdot \text{CLIP}(\vb*{c}) / \tau)}.
    \label{eq:loss}
\end{equation}
Then, the score function can be written as cosine similarity:
\begin{equation}
    score(\vb*{\vb*{g_i}, \text{CLIP}(\vb*{c})}) = \frac{\vb*{g_i} \cdot \vb*{\text{CLIP}(\vb*{c})}}{|\vb*{g_i}||\vb*{\text{CLIP}(\vb*{c})}|}.
\end{equation}
During inference, the prediction loop ends when the predictor matches the final state image.

\section{Experiments on TRANCE} \label{exp:exp_trance}

In this section, we first briefly introduce the experimental settings, and then show our experimental results on the three settings of TRANCE, i.e. Basic, Event, and View. We also conduct analyses to provide some insights about machines' ability of reasoning transformation.

We would like to test how well existing methods work on this new task. However, since the inputs and outputs of TVR are quite different from existing visual reasoning tasks, existing methods like~\cite{johnson2017inferring, hudson2018compositional} cannot be directly applied. Instead, we compare eight TranceNet variants as well as humans as the initial benchmark.

\textit{TranceNets.}
In the encoder part, we test two networks encoding images in the early fusion way, i.e.~Vanilla CNN and ResNet, combined with two fusion methods, i.e.~subtraction ($-$) and concatenation ($\oplus$), including CNN$_-$, CNN$_\oplus$, ResNet$_-$, ResNet$_\oplus$. And we test BCNN and DUDA as the encoders in the latter fusion way. The decoder of the first six models is GRU while the decoder of the last two models is transformer. The predictor is shared just as described in \cref{sec:model_trance}. We denote these models by their encoders' names suffixed with `G' and `T' to represent GRU decoder and transformer decoder respectively. For example, ResNet$\oplus$-G means the encoder is a ResNet feeding in concatenated image pairs and the decoder is a GRU. During training, teacher forcing~\cite{teacherforcing} is applied for faster convergence. More implementation details such as number of layers and kernel size can be found in the supplementary.

\textit{Human.} To compare with humans, for each of the three settings, we also collect the results of 100 samples in total. These results come from 10 CS Ph.D. candidates who are familiar with our problems and the testing system.

\begin{table}[t]
\centering
\caption{Results of ResNet$_{-}$-T trained using REINFORCE~\cite{williams1992simple} with different rewards on Event.}
\label{tab:rl}
\begin{small}
\begin{tabular}{lrrrr}
\toprule
Model & AD$\downarrow$ & AND$\downarrow$ & LAcc$\uparrow$ & Acc$\uparrow$ \\
\midrule
ResNet$_{-}$-T & 0.8389 & 0.2601 & 0.6865 & 0.6553 \\
\hspace{0.2em} + \textit{corr} & 0.7711 & 0.2367 & 0.7061 & 0.6729 \\
\hspace{0.2em} + \textit{dist} & 0.7741 & 0.2370 & 0.7065 & 0.6734 \\
\hspace{0.2em} + \textit{corr \& dist} & \textbf{0.7681} & \textbf{0.2354} & \textbf{0.7069} & \textbf{0.6740} \\
\bottomrule
\end{tabular}
\end{small}
\end{table}

\subsection{Results on Three Settings} \label{sec:main_res}

From the results of Basic in the left part of~\cref{tab:results}, we can see that all models perform quite well, in the sense that the performance gap between these models and the human is not very large. Now we compare these models, where the difference lies in the encoder, ResNet$_{-/\oplus}$-G performs better than BCNN-G and DUDA-G. Recall that CNN$_{-/\oplus}$ and ResNet$_{-/\oplus}$ are early fusion encoders while BCNN and DUDA are latter fusion encoders. We can conclude that the early fusion way is better than the latter fusion way on the Basic setting, as the parameter size of ResNet$_{-/\oplus}$, BCNN, and DUDA is similar. By looking closely to the fine-grained accuracy, we can see the way of encoding affect the ability to find the correct objects and values, while the ability to distinguish different attributes is almost the same.

The middle part of~\cref{tab:results} shows the experimental results of Event. The extremely big performance gap between models and humans suggests Event is very challenging for machines. The major reason is the answer space rises exponentially when the number of steps increases. In our experiments, the size of answer space is $\sum_{i=1}^4 (33 \times 10)^i$, about 11.86 billion. The performance (e.g.~Acc) gap between CNN$_{-/\oplus}$-G and ResNet$_{-/\oplus}$-G becomes even larger on Event compared with Basic, which suggests larger encoders have advantages in extracting sufficient features to decode transformation sequences. ResNet$_{-/\oplus}$-T performs better than ResNet$_{-/\oplus}$-G on ~5\% test samples, which shows the advantage of the transformer to the GRU.

We also employ reinforcement learning to train models. Specifically, the signals including the \textit{correctness} and the \textit{distance} of a prediction to the reference transformation can be easily obtained after a simulation. Therefore, these signals are able to be used as rewards in REINFORCE~\cite{williams1992simple} algorithm to further train ResNet$_{-}$-T models. \cref{tab:rl} shows that all three rewards significantly improve performance, and the difference among them is small.

The right part of~\cref{tab:results} shows the results of View. While humans are insensitive to view variations, the performances of all deep models drop sharply from Event to View according to $\Delta$Acc, from -0.0497 to -0.2119. Among these models, CNN models with fewer parameters drop more sharply while ResNet$_{-/\oplus}$-T have the least negative impacts, which shows larger models have positive benefits and the advantage of the transformer.

\begin{figure}[t]
    \centering
    \includegraphics[width=\linewidth]{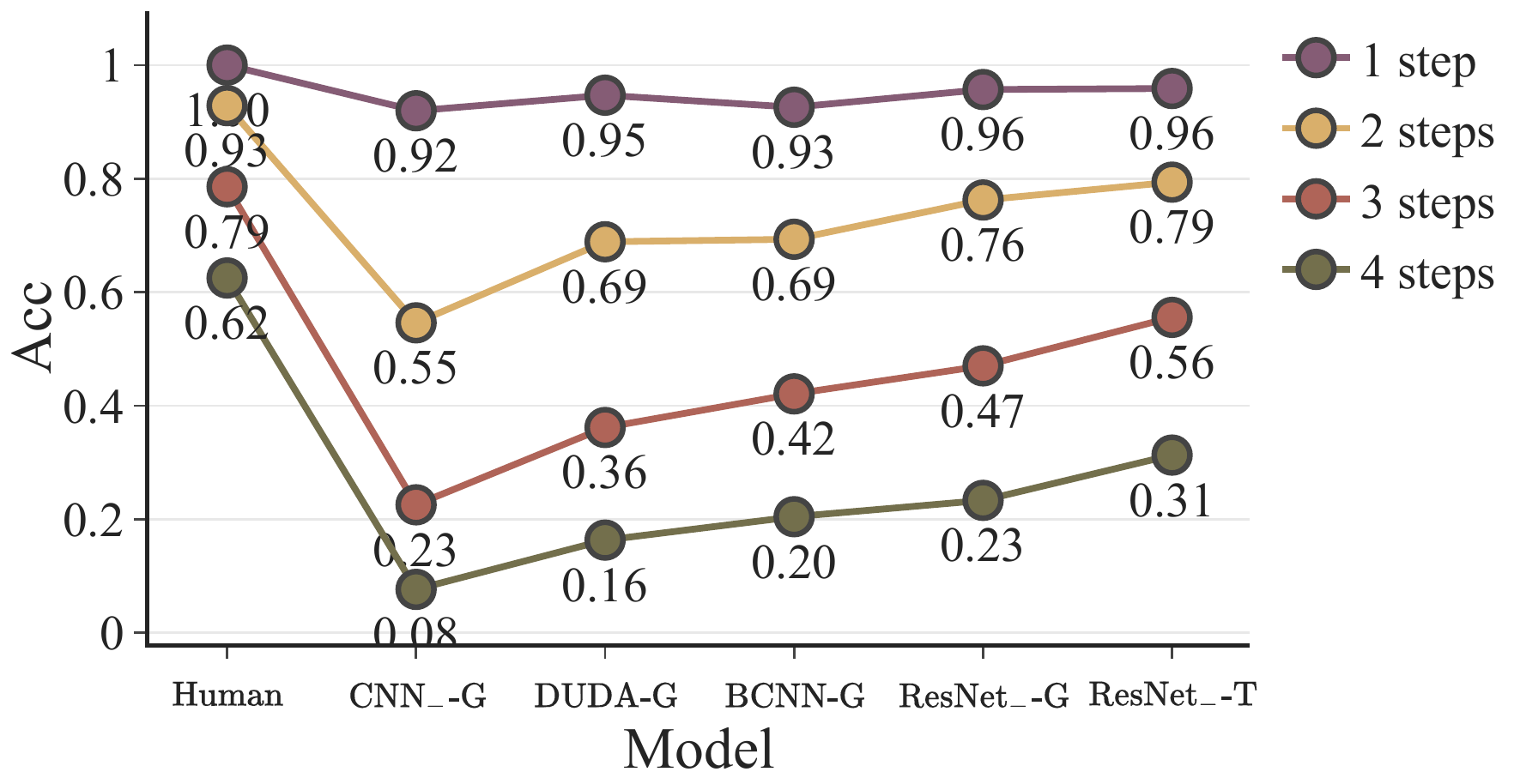}
    \caption{Results on Event with respect to different steps.}
    \label{fig:event}
\end{figure}

\subsection{Detailed Analysis on Event and View} \label{sec:analysis}

According to the above experimental results, models perform worse on Event and View. To understand the task more deeply and provide some insights for future model designs, we conduct a detailed analysis of two crucial factors of transformation, i.e. sequence length and order.

Firstly, we analyze the effect of transformation sequence length on Event, which is the major condition that differs from Basic. Specifically, we separate all test samples into four groups based on their lengths, i.e.~samples with $k$-step transformation $(k=1,2,3,4)$. Then we plot the Acc for each group in~\cref{fig:event}. From the results, both human and deep models work quite well when the length is short, e.g.~1. As the length increases, humans still capture complicated transformations very well. However, the performance of deep models declines sharply. Take CNN$_{-}$-G as an example, the performances for the four different groups are 92\%, 55\%, 23\%, and 8\%. These results indicate that future studies should focus more on how to tackle transformations with long steps. Another conclusion is that transformer is more advanced than GRU because of its higher ability of longer sequences modeling.

Then we analyze the effect of the order on Event, which is another important factor in this data. We collect results on order-sensitive samples. Specifically, we first build a subset of order-sensitive samples by testing each sample in the test set whether there exists a sequence permutation that prevents a successful transformation, caused by overlapping or moving out of the plane. We then test models on these samples, with 6.2\% \footnote{In another subset that only exists positional transformations, where ~25\% of them are order sensitive, the experimental results are similar.} samples from the test set and the result is shown in \cref{tab:order}. The metric EO is directly defined to measure the influence of order, LAcc and Acc are just listed for reference. From the results, we can see that EO of the human is zero. That is to say, once humans find all the correct atomic transformations, it is not hard to figure out the order. However, for all deep models, the EOs are larger than zero, which indicates a clear effect of the order on the reasoning process. In order to find out the extent of the effect, i.e., whether $0.0942\sim0.2050$ means a large deviation, we perform an experiment on 100 randomly selected order-sensitive samples. Specifically, we randomly permutate reference atomic transformations. As a result, the EO is 0.5008, which could be viewed as an upper bound of the order error. Therefore, the current deep models indeed have some ability to tackle the orders, but there still has a large room for improvement.

\begin{figure}[t]
    \centering
    \includegraphics[width=\linewidth]{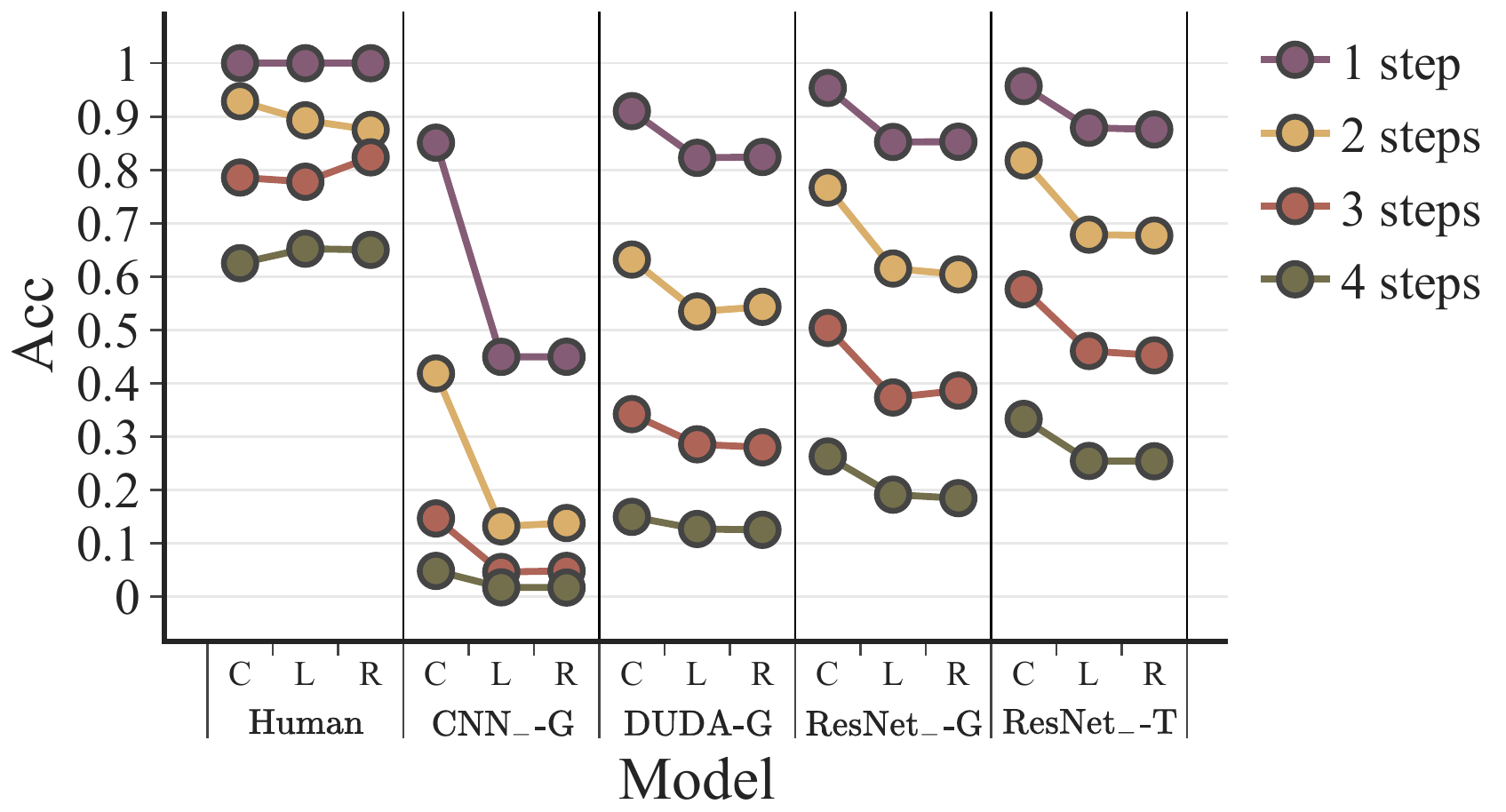}
    \caption{Results for different final views (\textbf{C}enter, \textbf{L}eft, \textbf{R}ight).}
    \label{fig:view_result}
\end{figure}

We finally analyze the effect of view variation. For each model, we provide the results of different final views, as shown in~\cref{fig:view_result}. Please note the results of CNN$_{\oplus}$-G, BCNN-G, ResNet$_{\oplus}$-G, and ResNet$_{\oplus}$-T are quite similar to CNN$_{-}$-G, DUDA-G, ResNet$_{-}$-G, and ResNet$_{-}$-T, so we just give the results from latter three typical models. The results of humans across different views change small, demonstrating human's powerful ability to adapt to different views. In some cases, humans perform even better when views are changed than unchanged. That is because humans usually spend more time solving problems when the view is altered, resulting in a decrease in the chance to make errors. Conversely, deep learning models share a similar trend that view variations will hurt performance. Among these models, CNN-G decreases the most, while DUDA-G shows its robustness. In conclusion, models with more parameters are more robust to view variations and feature-based interaction like the way used in DUDA-G is helpful.


\begin{table}[t]
\centering
\caption{Results on 6.2\% order sensitive samples from Event.}
\label{tab:order}
\begin{small}
\begin{tabular}{lrrr}
\toprule
Model & LAcc$\uparrow$ & Acc$\uparrow$ & EO$\downarrow$ \\
\midrule
Random (avg. of 100) & 1.0000 & 0.4992 & 0.5008 \\
\midrule
CNN$_{-}$-G & 0.1540 & 0.1395 & \textbf{0.0942} \\
DUDA-G & 0.1944 & 0.1613 & 0.1701 \\
BCNN-G & 0.2339 & 0.1935 & 0.1724 \\
ResNet$_{-}$-G & 0.3226 & 0.2565 & 0.2050 \\
ResNet$_{-}$-T & \textbf{0.3556} & \textbf{0.2911} & 0.1814 \\
\midrule
Human & 0.7273 & 0.7273 & 0.0000 \\
\bottomrule
\end{tabular}
\end{small}
\end{table}

\begin{table*}[t]
\begin{center}
\begin{small}

\caption{Model results on TRANCO (R and P are short for Recall and Precision).}
\label{tab:result_tranco}

\setlength{\tabcolsep}{0.45em}

\begin{tabular}{lrrrrrrrrrrrr}
\toprule

    \multirow{2}{*}{Model} & \multicolumn{6}{c}{Tiny Candidates (100)} & \multicolumn{6}{c}{Full Candidates (4918)} \\
     \cmidrule[.5pt](lr){2-7}  \cmidrule[.5pt](lr){8-13}
     & R $\uparrow$ & P $\uparrow$ & KTD $\downarrow$ & SD $\downarrow$ & NSD $\downarrow$ & EMR $\uparrow$ & R $\uparrow$ & P $\uparrow$ & KTD $\downarrow$ & SD $\downarrow$ & NSD $\downarrow$ &   EMR $\uparrow$ \\
\midrule
Random & 0.0316 & 0.0409 & 0.9958 & 1.9364 & 0.6845 & 0.0000 & 0.0005 & 0.0006 & 1.0000 & 1.9888 & 0.7093 & 0.0000 \\
\midrule

   RN101-G &           0.5733 &  \textbf{0.8633} &               0.4540 &              1.5503 &               0.4002 &             0.1524 &          0.3374 &          0.4865 &           0.7489 &          1.8147 &           0.5509 &          0.0490 \\
ViT-B/32-T &  \textbf{0.7549} &           0.8434 &               0.2981 &              1.4154 &               0.4248 &             0.1986 & \textbf{0.4890} &          0.5205 &           0.5982 &          1.7399 &           0.5709 & \textbf{0.0881} \\
ViT-B/16-T &           0.7416 &           0.8420 &               0.2912 &              1.4413 &               0.4343 &             0.1860 &          0.4883 & \textbf{0.5394} &           0.5933 &          1.6007 &           0.5148 &          0.0811 \\
   RN101-T &           0.7188 &           0.8620 &      \textbf{0.2545} &     \textbf{1.1154} &      \textbf{0.2908} &    \textbf{0.2161} &          0.4598 &          0.5006 &  \textbf{0.5865} & \textbf{1.2832} &  \textbf{0.3976} &          0.0727 \\

\bottomrule
\end{tabular}

\end{small}
\end{center}
\label{tab:backbone}
\end{table*}



\section{Experiments on TRANCO} \label{sec:exp_tranco}

The previous section has analyzed the experimental results of the synthetic dataset TRANCE. The following section will move to analyze how models perform on real data. Similar to the previous section, the experimental setting is first briefly introduced, then we show the analysis of results.

In terms of comparing baselines, we first set a random baseline to provide the lower bound of the performance as a reference. And we compare five TrancoNet models to set the initial benchmark for TRANCO.

\textit{Random.} First, the total number of steps $n$ is randomly selected from 2 to 7. Next, $n$ non-repeating atomic transformations are sequentially and randomly sampled from the candidate set as the prediction.

\textit{TrancoNets.}  In the encoder part, we consider three types of encoder borrowed from CLIP~\cite{Radford2021LearningTV}, including RN101, ViT-B/16, and ViT-B/32. The input images are encoded in the latter fusion way. In the decoder part, except for the transformer decoder described in \cref{sec:model_tranco}, the GRU decoder is also compared. These models are denoted by their encoders' names suffixed with `G' or `T', indicating GRU and transformer respectively. During training, it is computationally expensive if all available video clips in the training set are included in the candidate. Therefore, for each sample, we randomly select negative atomic transformations from other training samples, to constitute a candidate set size of 20, which is a trade-off between performance and resource consumption. Further analysis of the candidate set size and more implementation details of models are included in the supplementary material.

During the evaluation, in addition to the full test candidates, which contain 4918 atomic transformations (video clips), we also construct a tiny candidate of size 100 for each sample. This can help us to learn how candidate size affects the model's performance. The results on tiny candidates are suffixed with `@100', e.g. EMR@100.

\subsection{Results on TRANCO}
\label{sec:results_tranco}

\cref{tab:result_tranco} show the performance of five models on two sizes of candidates. From the table, we can see that EMR@100 of the random baseline is exactly zero. This is because the transformation space is large, which is a combination of different atomic transformations with different orders. Given such a huge space, it is almost impossible to find a correct answer by finding random atomic transformations and assigning a random order. Another comparison is between the results of ResNet$_{\oplus}$-G on TRANCE and the results of RN101-G here. While RN101-G has more parameters than ResNet$_{\oplus}$-G, and is pretrained, the EMR on TRANCO (0.0490) is much lower than Acc on View (0.5425). These results show TRANCO is hard, much more difficult than TRANCE. Next, by comparing the left part of the table with the right part, we can find that compared with EMR on tiny candidates, EMR of all models on full candidates drops by more than 60 percent, which suggests that the high diversity of atomic transformations is one reason that TRANCO is difficult. Finally, the results between transformer based models and GRU based models show transformer performs better on reasoning transformations. The large gap in recall and KTD indicates that transformer is more outstanding in finding complete atomic transformations and capturing the order.

\begin{figure}[t]
    \centering
    \includegraphics[width=\linewidth]{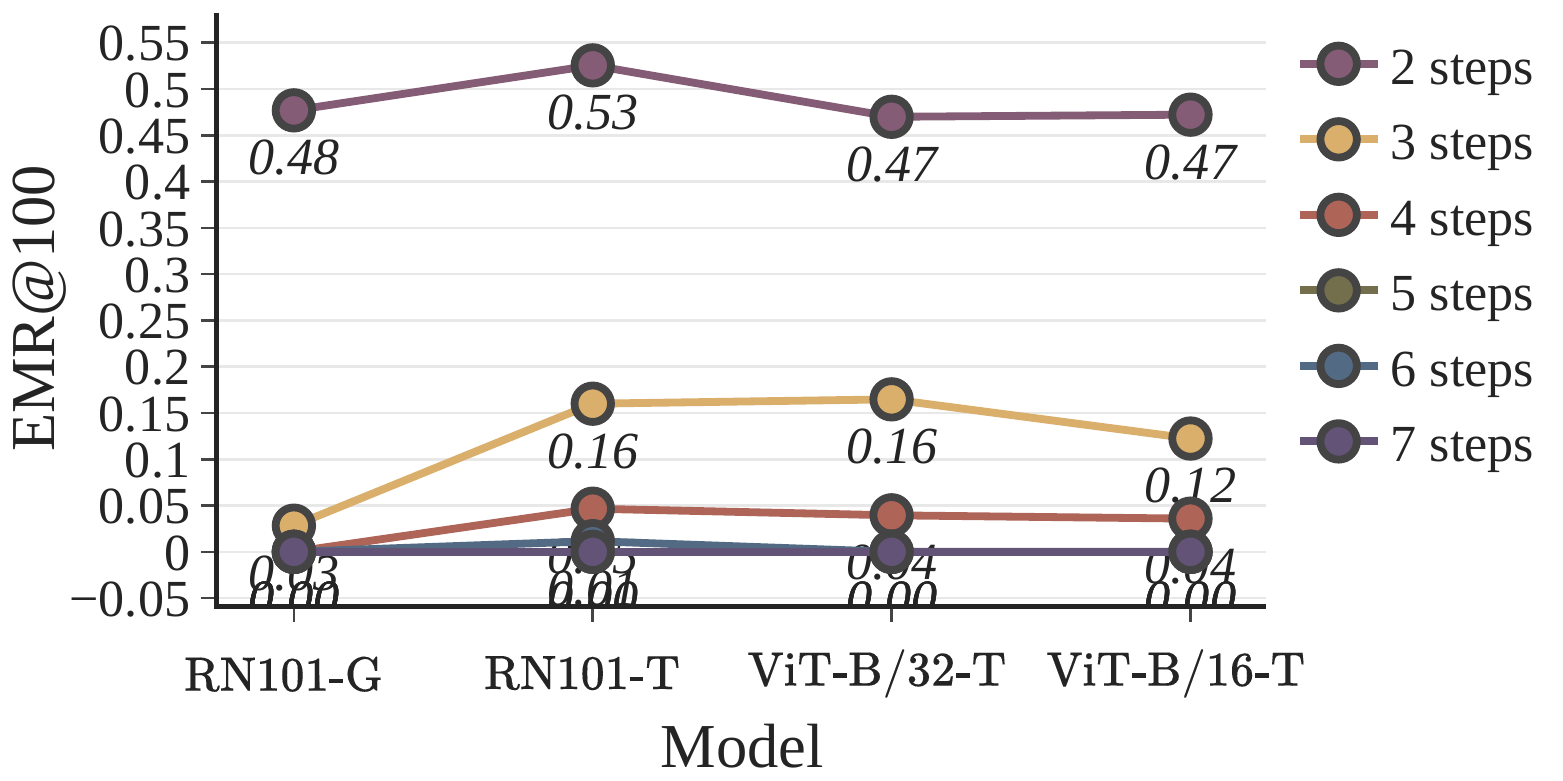}
    \caption{Results on TRANCO with respect to different steps.}
    \label{fig:tranco_step}
\end{figure}

\begin{table}[t]
\centering
\small

\caption{Results on TRANCO with respect to different pretraining strategies.}
\label{tab:pretrain}
\begin{tabular}{lrr}
\toprule
    Pretrain strategy & EMR@100 $\uparrow$ &   EMR $\uparrow$ \\
\midrule
         from scratch &             0.1210 &           0.0378 \\
pretrain w/o finetune &   \textbf{0.1986} & \textbf{0.0881} \\
 pretrain w/ finetune &             0.1965 &           0.0748 \\
\bottomrule
\end{tabular}
\end{table}

\subsection{Detailed Analysis on TRANCO}

As previously analyzed on TRANCE, sequence length and order are two important factors for transformation reasoning. In this section, we analyze the impact of sequence length again. However, the order is not able to be further analyzed since evaluating order on real data is not convenient. Instead, we will analyze how pretrained CLIP matters, since real data requires additional recognition ability and pretrained CLIP is expected to do well.

We first analyze how transformation sequence length affects the model's performance. The results are shown in \cref{fig:tranco_step}. The length of transformation is ranged from 2 to 7 on TRANCO. From the results, we can see that models answer half of 2-step samples correctly. However, the EMR@100 drops sharply when the length is larger than 2 and becomes zero when the length is larger than 4. These results prove the previous findings on TRANCE that transformations with more steps are difficult and should be focused on in future studies. Another finding is transformer indeed performs better than GRU on longer-length transformations due to its outstanding ability on capturing long-range dependence.

Another important problem is how pretrained CLIP benefits models. Therefore, we compare three different strategies for training ViT-B/32-T and the results are shown in \cref{tab:pretrain}. We can see that models initialized with pretrained weights perform much better than models trained from scratch, improving about 60\% on EMR@100 and 100\% on EMR. During training, we also observe that models initialized with pre-trained weights converge much faster. All these results suggest pretrained weights from CLIP indeed benefit the transformation reasoning, with its strong ability on extracting semantic meaningful representation. However, the performance drops slightly when the pretrained weights are further tuned. By jointly analyzing the EMR curve during training, we find tuning pretrained weights results in overfitting while fixing pretrained weights does not. We believe the small training set does not support further tuning a better feature extractor, therefore the pretrained weights are fixed in all other experiments on TRANCO.




\section{Discussion: from TRANCE to TRANCO}

\begin{figure}[t]
    \centering
    \includegraphics[width=\linewidth]{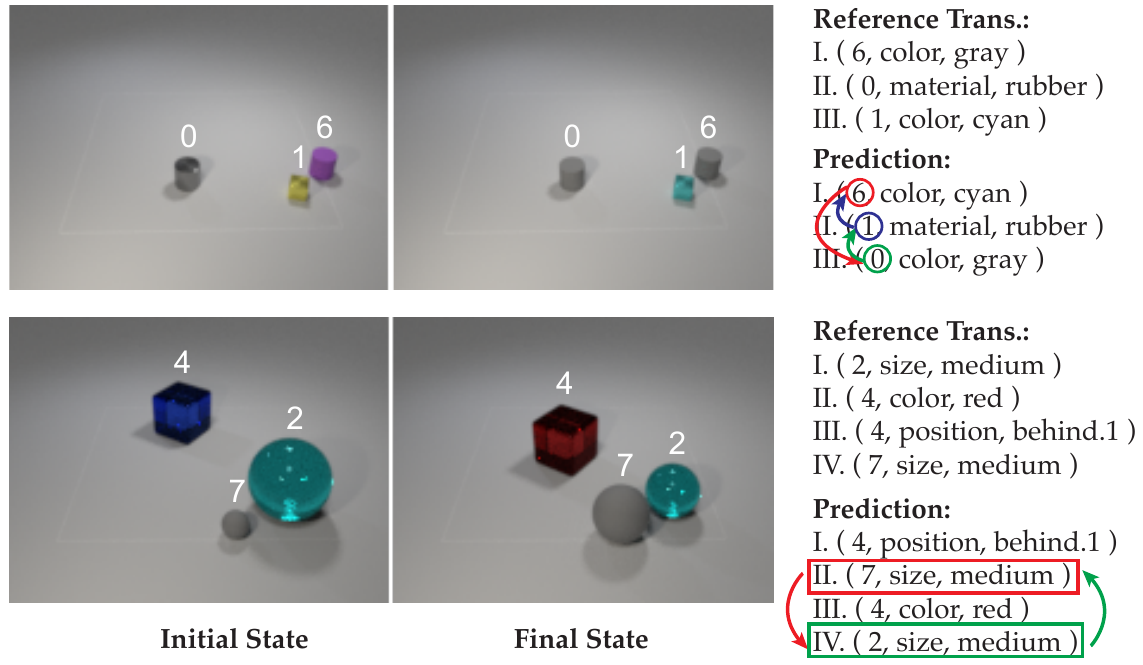}
    \caption{Typical failure cases in TRANCE. In the first case, the model finds all objects and actions but they are mismatched. In the second case, the model finds all atomic transformations, but two of them are in reverse order.}
    \label{fig:failure_cases}
\end{figure}

From the experimental results on TRANCE (e.g. Event) and TRANCO, there are some similarities and differences between the synthetic and real settings. The biggest similarity is that transformations with more steps are more difficult to be reasoned correctly, according to \cref{fig:event} and \cref{fig:tranco_step}. With a deeper analysis of the failure cases from the two datasets, we find the types of mistakes are slightly different. In TRANCE, even in failure cases, models are able to find most objects and actions but may fail to match the action to the correct object or find a correct order, as shown in ~\cref{fig:failure_cases}. While in TRANCO, models even fail to find all correct transformations from candidates most of the time, let alone the right order. This is mainly due to the different characteristics of the two problems. Objects and their attributes are simple in TRANCE but are significantly diverse in TRANCO. Therefore, the requirement for image recognition ability is higher on TRANCO. This is why we empirically found pretrained ResNet has little positive effects on TRANCE but pretrained image encoders such as CLIP make a huge difference (\cref{tab:pretrain}) on TRANCO. However, both datasets require context reasoning ability to generate the correct sequence of transformations, especially when the number of steps is large. Transformer is known to be good at modeling long range dependencies, and this is why it performs better than GRU on both problems. From these two observations, we believe that improving visual transformation reasoning is primarily a matter of finding models with greater abilities of image recognition and contextual reasoning, to make models robust even when reasoning transformations with many steps.

\section{Conclusion}

To tackle the problem that most existing visual reasoning tasks are solely defined in static settings and cannot well capture the dynamics between states, we propose a new visual reasoning paradigm, namely transformation driven visual reasoning (TVR). Given the initial and final states, the target is to infer the corresponding sequence of atomic transformations, while the atomic transformation is represented by a triplet (object, attribute, value) or a video clip. In this paper, as an example, we use CLEVR to construct a new synthetic data, namely TRANCE, which includes three different levels of settings, i.e.~Basic for single-step transformation, Event for multi-step transformation, and View for multi-step transformation with variant views. We also construct a real dataset called TRANCO to test reasoning ``open-world'' transformations. To study the effectiveness of existing SOTA reasoning techniques, we propose a human-inspired reasoning framework named TranNet. The experimental results show that our best model works well on Basic, while still having difficulties solving Event, View, and more difficult TRANCO. Specifically, the difficult point of Event is to find all atomic transformations and arrange them with a feasible order, especially when the length of the sequence is large. The view variations in View bring great challenges to these models, but have little impact on humans. While for TRANCO, it brings extra challenges with massive diverse atomic transformations.


\ifCLASSOPTIONcompsoc
  \section*{Acknowledgments}
\else
  \section*{Acknowledgment}
\fi


This work was supported by the National Key R\&D Program of China under Grant 2022YFB3103704, in part by the National Natural Science Foundation of China (NSFC) under Grant 62276248, and in part by Beijing Academy of Artificial Intelligence (BAAI) under Grant BAAI2020ZJ0303.

\ifCLASSOPTIONcaptionsoff
  \newpage
\fi



\bibliographystyle{IEEEtran}
\bibliography{IEEEabrv, main}

\begin{thebibliography}{10}
\providecommand{\url}[1]{#1}
\csname url@samestyle\endcsname
\providecommand{\newblock}{\relax}
\providecommand{\bibinfo}[2]{#2}
\providecommand{\BIBentrySTDinterwordspacing}{\spaceskip=0pt\relax}
\providecommand{\BIBentryALTinterwordstretchfactor}{4}
\providecommand{\BIBentryALTinterwordspacing}{\spaceskip=\fontdimen2\font plus
\BIBentryALTinterwordstretchfactor\fontdimen3\font minus
  \fontdimen4\font\relax}
\providecommand{\BIBforeignlanguage}[2]{{%
\expandafter\ifx\csname l@#1\endcsname\relax
\typeout{** WARNING: IEEEtran.bst: No hyphenation pattern has been}%
\typeout{** loaded for the language `#1'. Using the pattern for}%
\typeout{** the default language instead.}%
\else
\language=\csname l@#1\endcsname
\fi
#2}}
\providecommand{\BIBdecl}{\relax}
\BIBdecl

\bibitem{johnson2017clevr}
J.~Johnson, B.~Hariharan, L.~van~der Maaten, L.~Fei{-}Fei, C.~L. Zitnick, and
  R.~B. Girshick, ``{CLEVR:} {A} diagnostic dataset for compositional language
  and elementary visual reasoning,'' in \emph{{IEEE Conf. Comput. Vis. Pattern
  Recog.}}, 2017, pp. 1988--1997.

\bibitem{suhr2017corpus}
A.~Suhr, M.~Lewis, J.~Yeh, and Y.~Artzi, ``A corpus of natural language for
  visual reasoning,'' in \emph{{Proc. Annu. Meet. Assoc. Comput. Linguistics}},
  2017, pp. 217--223.

\bibitem{suhr2019corpus}
A.~Suhr, S.~Zhou, A.~Zhang, I.~Zhang, H.~Bai, and Y.~Artzi, ``A corpus for
  reasoning about natural language grounded in photographs,'' in \emph{{Proc.
  Annu. Meet. Assoc. Comput. Linguistics}}, 2019, pp. 6418--6428.

\bibitem{zellers2019vcr}
R.~Zellers, Y.~Bisk, A.~Farhadi, and Y.~Choi, ``From recognition to cognition:
  Visual commonsense reasoning,'' in \emph{{IEEE Conf. Comput. Vis. Pattern
  Recog.}}, 2019, pp. 6720--6731.

\bibitem{piaget1977role}
J.~Piaget, ``The role of action in the development of thinking,'' in
  \emph{Knowledge and Development}, 1977, pp. 17--42.

\bibitem{huang2016visual}
T.-H.~K. Huang, F.~Ferraro, N.~Mostafazadeh, I.~Misra, A.~Agrawal, J.~Devlin,
  R.~Girshick, X.~He, P.~Kohli, D.~Batra, C.~L. Zitnick, D.~Parikh,
  L.~Vanderwende, M.~Galley, and M.~Mitchell, ``Visual storytelling,'' in
  \emph{{Proc. Conf. North Amer. Chapter Assoc. Comput. Linguistics: Hum. Lang.
  Technol.}}, 2016, pp. 1233--1239.

\bibitem{park2020visualcomet}
J.~S. Park, C.~Bhagavatula, R.~Mottaghi, A.~Farhadi, and Y.~Choi,
  ``{VisualCOMET}: Reasoning about the dynamic context of a still image,'' in
  \emph{{Eur. Conf. Comput. Vis.}}, 2020, pp. 508--524.

\bibitem{Bendale2015TowardsOW}
A.~Bendale and T.~E. Boult, ``Towards open world recognition,'' in \emph{{IEEE
  Conf. Comput. Vis. Pattern Recog.}}, 2015, pp. 1893--1902.

\bibitem{johnson2017inferring}
J.~Johnson, B.~Hariharan, L.~van~der Maaten, J.~Hoffman, L.~Fei{-}Fei, C.~L.
  Zitnick, and R.~B. Girshick, ``Inferring and executing programs for visual
  reasoning,'' in \emph{{Int. Conf. Comput. Vis.}}, 2017, pp. 3008--3017.

\bibitem{hudson2018compositional}
D.~A. Hudson and C.~D. Manning, ``Compositional attention networks for machine
  reasoning,'' in \emph{{Int. Conf. Learn. Represent.}}, 2018.

\bibitem{he2016deep}
K.~He, X.~Zhang, S.~Ren, and J.~Sun, ``Deep residual learning for image
  recognition,'' in \emph{{IEEE Conf. Comput. Vis. Pattern Recog.}}, 2016, pp.
  770--778.

\bibitem{lin2015bilinear}
T.~Lin, A.~RoyChowdhury, and S.~Maji, ``Bilinear {CNN} models for fine-grained
  visual recognition,'' in \emph{{Int. Conf. Comput. Vis.}}, 2015, pp.
  1449--1457.

\bibitem{Park_2019_ICCV}
D.~H. Park, T.~Darrell, and A.~Rohrbach, ``Robust change captioning,'' in
  \emph{{Int. Conf. Comput. Vis.}}, 2019, pp. 4623--4632.

\bibitem{Radford2021LearningTV}
A.~Radford, J.~W. Kim, C.~Hallacy, A.~Ramesh, G.~Goh, S.~Agarwal, G.~Sastry,
  A.~Askell, P.~Mishkin, J.~Clark, G.~Krueger, and I.~Sutskever, ``Learning
  transferable visual models from natural language supervision,'' in
  \emph{{Int. Conf. Mach. Learn.}}, 2021, pp. 8748--8763.

\bibitem{cho2014learning}
K.~Cho, B.~van Merri{\"e}nboer, C.~Gulcehre, D.~Bahdanau, F.~Bougares,
  H.~Schwenk, and Y.~Bengio, ``Learning phrase representations using {RNN}
  encoder{--}decoder for statistical machine translation,'' in \emph{Proc. of
  EMNLP}, 2014, pp. 1724--1734.

\bibitem{vaswani2017attention}
A.~Vaswani, N.~Shazeer, N.~Parmar, J.~Uszkoreit, L.~Jones, A.~N. Gomez,
  L.~Kaiser, and I.~Polosukhin, ``Attention is all you need,'' in \emph{{Adv.
  Neural Inform. Process. Syst.}}, 2017, pp. 5998--6008.

\bibitem{VQA}
S.~Antol, A.~Agrawal, J.~Lu, M.~Mitchell, D.~Batra, C.~L. Zitnick, and
  D.~Parikh, ``{VQA:} visual question answering,'' in \emph{{Int. Conf. Comput.
  Vis.}}, 2015, pp. 2425--2433.

\bibitem{balanced_binary_vqa}
P.~Zhang, Y.~Goyal, D.~Summers{-}Stay, D.~Batra, and D.~Parikh, ``Yin and yang:
  Balancing and answering binary visual questions,'' in \emph{{IEEE Conf.
  Comput. Vis. Pattern Recog.}}, 2016, pp. 5014--5022.

\bibitem{goyal2017making}
Y.~Goyal, T.~Khot, D.~Summers{-}Stay, D.~Batra, and D.~Parikh, ``Making the {V}
  in {VQA} matter: Elevating the role of image understanding in visual question
  answering,'' in \emph{{IEEE Conf. Comput. Vis. Pattern Recog.}}, 2017, pp.
  6325--6334.

\bibitem{zhu2016visual7w}
Y.~Zhu, O.~Groth, M.~S. Bernstein, and L.~Fei{-}Fei, ``Visual7w: Grounded
  question answering in images,'' in \emph{{IEEE Conf. Comput. Vis. Pattern
  Recog.}}, 2016, pp. 4995--5004.

\bibitem{marino2019ok}
K.~Marino, M.~Rastegari, A.~Farhadi, and R.~Mottaghi, ``{OK-VQA:} {A} visual
  question answering benchmark requiring external knowledge,'' in \emph{{IEEE
  Conf. Comput. Vis. Pattern Recog.}}, 2019, pp. 3195--3204.

\bibitem{Hudson_2019_CVPR}
D.~A. Hudson and C.~D. Manning, ``{GQA:} {A} new dataset for real-world visual
  reasoning and compositional question answering,'' in \emph{{IEEE Conf.
  Comput. Vis. Pattern Recog.}}, 2019, pp. 6700--6709.

\bibitem{xie2018visual}
N.~Xie, F.~Lai, and D.~a. Doran, ``Visual entailment task for visually-grounded
  languag,'' \emph{arXiv:1811.10582}, 2018.

\bibitem{xie2019visual}
N.~Xie, F.~Lai, D.~Doran, and A.~Kadav, ``Visual entailment: A novel task for
  fine-grained image understanding,'' \emph{arXiv:1901.06706}, 2019.

\bibitem{wang2018fvqa}
P.~Wang, Q.~Wu, C.~Shen, A.~Dick, and A.~van~den Hengel, ``{FVQA}: Fact-based
  visual question answering,'' \emph{{IEEE Trans. Pattern Anal. Mach.
  Intell.}}, pp. 2413--2427, 2018.

\bibitem{girdhar2020cater}
R.~Girdhar and D.~Ramanan, ``{CATER: A diagnostic dataset for Compositional
  Actions and TEmporal Reasoning},'' in \emph{{Int. Conf. Learn. Represent.}},
  2020.

\bibitem{yi2020clevrer}
K.~Yi, C.~Gan, Y.~Li, P.~Kohli, J.~Wu, A.~Torralba, and J.~B. Tenenbaum,
  ``{CLEVRER:} collision events for video representation and reasoning,'' in
  \emph{{Int. Conf. Learn. Represent.}}, 2020.

\bibitem{bakhtin2019phyre}
A.~Bakhtin, L.~van~der Maaten, J.~Johnson, L.~Gustafson, and R.~B. Girshick,
  ``{PHYRE:} {A} new benchmark for physical reasoning,'' in \emph{{Adv. Neural
  Inform. Process. Syst.}}, 2019, pp. 5083--5094.

\bibitem{Baradel_2020_ICLR_cophy}
F.~Baradel, N.~Neverova, J.~Mille, G.~Mori, and C.~Wolf, ``Cophy:
  Counterfactual learning of physical dynamics,'' in \emph{{Int. Conf. Learn.
  Represent.}}, 2020.

\bibitem{isola2015discovering}
P.~Isola, J.~J. Lim, and E.~H. Adelson, ``Discovering states and
  transformations in image collections,'' in \emph{{IEEE Conf. Comput. Vis.
  Pattern Recog.}}, 2015, pp. 1383--1391.

\bibitem{nagarajan2018attributes}
T.~Nagarajan and K.~Grauman, ``Attributes as operators: factorizing unseen
  attribute-object compositions,'' in \emph{{Eur. Conf. Comput. Vis.}}, 2018,
  pp. 169--185.

\bibitem{li2020symmetry}
Y.~Li, Y.~Xu, X.~Mao, and C.~Lu, ``Symmetry and group in attribute-object
  compositions,'' in \emph{{IEEE Conf. Comput. Vis. Pattern Recog.}}, 2020, pp.
  11\,313--11\,322.

\bibitem{fathi2013modeling}
A.~Fathi and J.~M. Rehg, ``Modeling actions through state changes,'' in
  \emph{{IEEE Conf. Comput. Vis. Pattern Recog.}}, 2013, pp. 2579--2586.

\bibitem{wang2016actions}
X.~Wang, A.~Farhadi, and A.~Gupta, ``Actions\~{} transformations,'' in
  \emph{{IEEE Conf. Comput. Vis. Pattern Recog.}}, 2016, pp. 2658--2667.

\bibitem{liu2017jointly}
Y.~Liu, P.~Wei, and S.~Zhu, ``Jointly recognizing object fluents and tasks in
  egocentric videos,'' in \emph{{Int. Conf. Comput. Vis.}}, 2017, pp.
  2943--2951.

\bibitem{alayrac2017joint}
J.~Alayrac, J.~Sivic, I.~Laptev, and S.~Lacoste{-}Julien, ``Joint discovery of
  object states and manipulation actions,'' in \emph{{Int. Conf. Comput.
  Vis.}}, 2017, pp. 2146--2155.

\bibitem{zhuo2019explainable}
T.~Zhuo, Z.~Cheng, P.~Zhang, Y.~Wong, and M.~S. Kankanhalli, ``Explainable
  video action reasoning via prior knowledge and state transitions,'' in
  \emph{{ACM Int. Conf. Multimedia}}, 2019, pp. 521--529.

\bibitem{Chang2020ProcedurePI}
C.~Chang, D.-A. Huang, D.~Xu, E.~Adeli, L.~Fei-Fei, and J.~C. Niebles,
  ``Procedure planning in instructional videos,'' in \emph{{Eur. Conf. Comput.
  Vis.}}, 2020.

\bibitem{blender2016blender}
\BIBentryALTinterwordspacing
B.~O. Community, \emph{Blender - a 3D modelling and rendering package}, Blender
  Foundation, Stichting Blender Foundation, Amsterdam, 2018. [Online].
  Available: \url{http://www.blender.org}
\BIBentrySTDinterwordspacing

\bibitem{liu2019clevr}
R.~Liu, C.~Liu, Y.~Bai, and A.~L. Yuille, ``Clevr-ref+: Diagnosing visual
  reasoning with referring expressions,'' in \emph{{IEEE Conf. Comput. Vis.
  Pattern Recog.}}, 2019, pp. 4185--4194.

\bibitem{jin2013spoid}
J.-H. Jin, H.~J. Shin, and J.-J. Choi, ``Spoid: a system to produce
  spot-the-difference puzzle images with difficulty,'' \emph{The Visual
  Computer}, pp. 481--489, 2013.

\bibitem{Tang2019COINAL}
Y.~Tang, D.~Ding, Y.~Rao, Y.~Zheng, D.~Zhang, L.~Zhao, J.~Lu, and J.~Zhou,
  ``{COIN:} {A} large-scale dataset for comprehensive instructional video
  analysis,'' in \emph{{IEEE Conf. Comput. Vis. Pattern Recog.}}, 2019, pp.
  1207--1216.

\bibitem{kendall1938}
M.~G. Kendall, ``A {{New Measure}} of {{Rank Correlation}},''
  \emph{Biometrika}, pp. 81--93, 1938.

\bibitem{osinga2007science}
F.~P. Osinga, \emph{Science, strategy and war: The strategic theory of John
  Boyd}.\hskip 1em plus 0.5em minus 0.4em\relax Routledge, 2007.

\bibitem{dosovitskiy2021an}
A.~Dosovitskiy, L.~Beyer, A.~Kolesnikov, D.~Weissenborn, X.~Zhai,
  T.~Unterthiner, M.~Dehghani, M.~Minderer, G.~Heigold, S.~Gelly, J.~Uszkoreit,
  and N.~Houlsby, ``An image is worth 16x16 words: Transformers for image
  recognition at scale,'' in \emph{{Int. Conf. Learn. Represent.}}, 2021.

\bibitem{PortilloQuintero2021ASF}
J.~A. Portillo-Quintero, J.~C. Ortiz-Bayliss, and H.~Terashima-Mar\'{\i}n, ``A
  straightforward framework for video retrieval using clip,'' in \emph{{Pattern
  Recogn.}}, 2021, pp. 3--12.

\bibitem{teacherforcing}
R.~J. {Williams} and D.~{Zipser}, ``A learning algorithm for continually
  running fully recurrent neural networks,'' \emph{Neural Computation}, pp.
  270--280, 1989.

\bibitem{williams1992simple}
R.~J. Williams, ``Simple statistical gradient-following algorithms for
  connectionist reinforcement learning,'' \emph{Machine learning}, pp.
  229--256, 1992.

\bibitem{NEURIPS2019_9015}
A.~Paszke, S.~Gross, F.~Massa, A.~Lerer, J.~Bradbury, G.~Chanan, T.~Killeen,
  Z.~Lin, N.~Gimelshein, L.~Antiga, A.~Desmaison, A.~Kopf, E.~Yang, Z.~DeVito,
  M.~Raison, A.~Tejani, S.~Chilamkurthy, B.~Steiner, L.~Fang, J.~Bai, and
  S.~Chintala, ``Pytorch: An imperative style, high-performance deep learning
  library,'' in \emph{{Adv. Neural Inform. Process. Syst.}}, 2019, pp.
  8024--8035.

\bibitem{simonyan2014very}
K.~Simonyan and A.~Zisserman, ``Very deep convolutional networks for
  large-scale image recognition,'' in \emph{{Int. Conf. Learn. Represent.}},
  2015.

\bibitem{DBLP:journals/corr/KingmaB14}
D.~P. Kingma and J.~Ba, ``Adam: {A} method for stochastic optimization,'' in
  \emph{{Int. Conf. Learn. Represent.}}, 2015.

\bibitem{krizhevsky2012imagenet}
A.~Krizhevsky, I.~Sutskever, and G.~E. Hinton, ``Imagenet classification with
  deep convolutional neural networks,'' in \emph{{Adv. Neural Inform. Process.
  Syst.}}, 2012, pp. 1106--1114.

\bibitem{Loshchilov2019DecoupledWD}
I.~Loshchilov and F.~Hutter, ``Decoupled weight decay regularization,'' in
  \emph{{Int. Conf. Learn. Represent.}}, 2019.

\bibitem{wang2016_TemporalSegmentNetworks}
L.~Wang, Y.~Xiong, Z.~Wang, Y.~Qiao, D.~Lin, X.~Tang, and L.~{Val Gool},
  ``Temporal segment networks: Towards good practices for deep action
  recognition,'' in \emph{{Eur. Conf. Comput. Vis.}}, 2016.

\end{thebibliography}
%



%
\vskip -2\baselineskip plus -1fil

\balance

\begin{IEEEbiography}[{\includegraphics[width=1in,height=1.25in,clip,keepaspectratio]{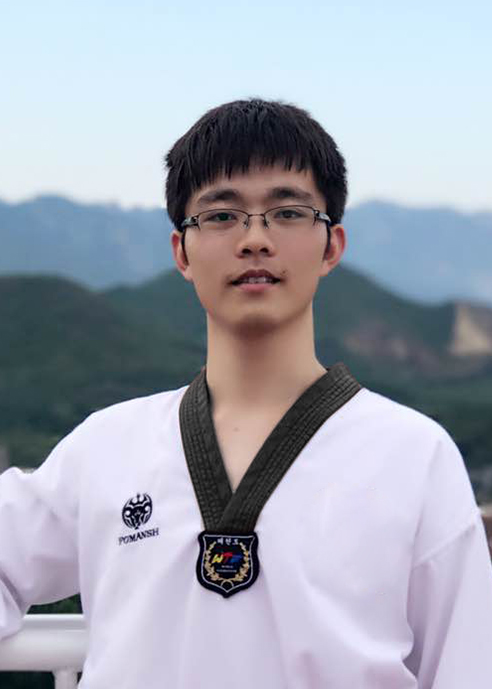}}]{Xin Hong} received the BEng degree in Software Engineering from Beijing University of Technology. He is currently working toward the PhD degree from University of Chinese Academy of Sciences. His main research interests include visual reasoning, multi-modal learning and image inpainting.
\end{IEEEbiography}

\vskip -2\baselineskip plus -1fil

\begin{IEEEbiography}[{\includegraphics[width=1in,height=1.25in,clip,keepaspectratio]{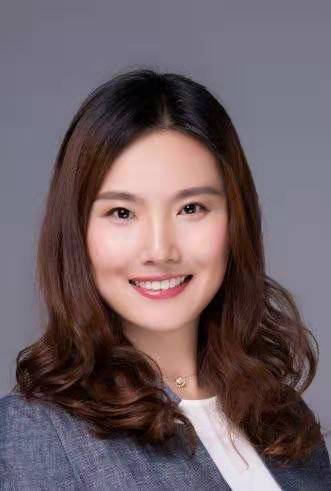}}]{Yanyan Lan} received the PhD degree in Probability and Statistics from the Institute of Applied Mathematics, Academy of Mathematics and System Sciences, Chinese Academy of Sciences (CAS). She is currently a professor at the Institute of AI Industrial Research, Tsinghua University. Her research interests include AI+Healthcare, information retrieval, machine learning and natural language processing.
\end{IEEEbiography}

\vskip -2\baselineskip plus -1fil


\begin{IEEEbiography}[{\includegraphics[width=1in,height=1.25in,clip,keepaspectratio]{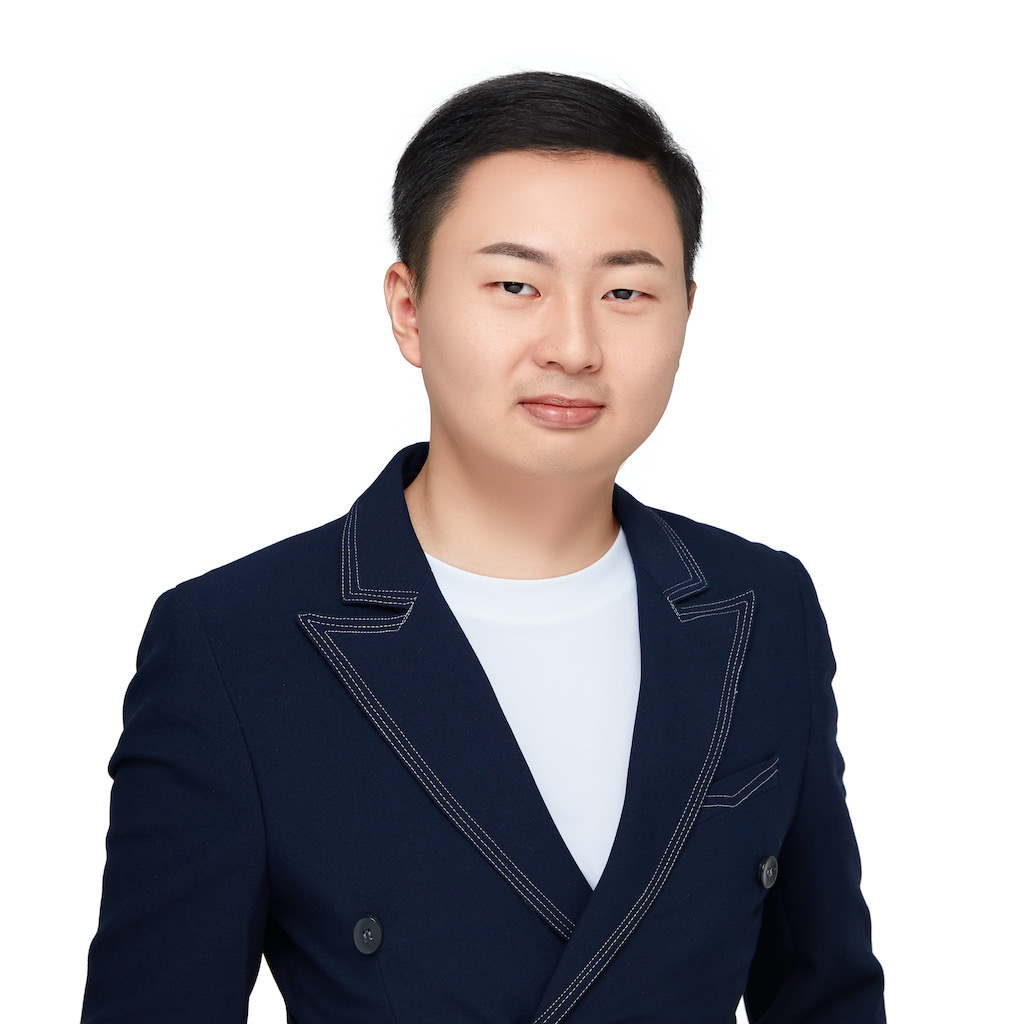}}]{Liang Pang} received the PhD degree in Computer Science from the University of Chinese Academy of Sciences. He is currently an associate researcher at Data Intelligence System Research Center, Institute of Computing Technology (ICT), Chinese Academy of Sciences (CAS). His research interests include natural language generation and information retrieval.
\end{IEEEbiography}

\vskip -2\baselineskip plus -1fil

\begin{IEEEbiography}[{\includegraphics[width=1in,height=1.25in,clip,keepaspectratio]{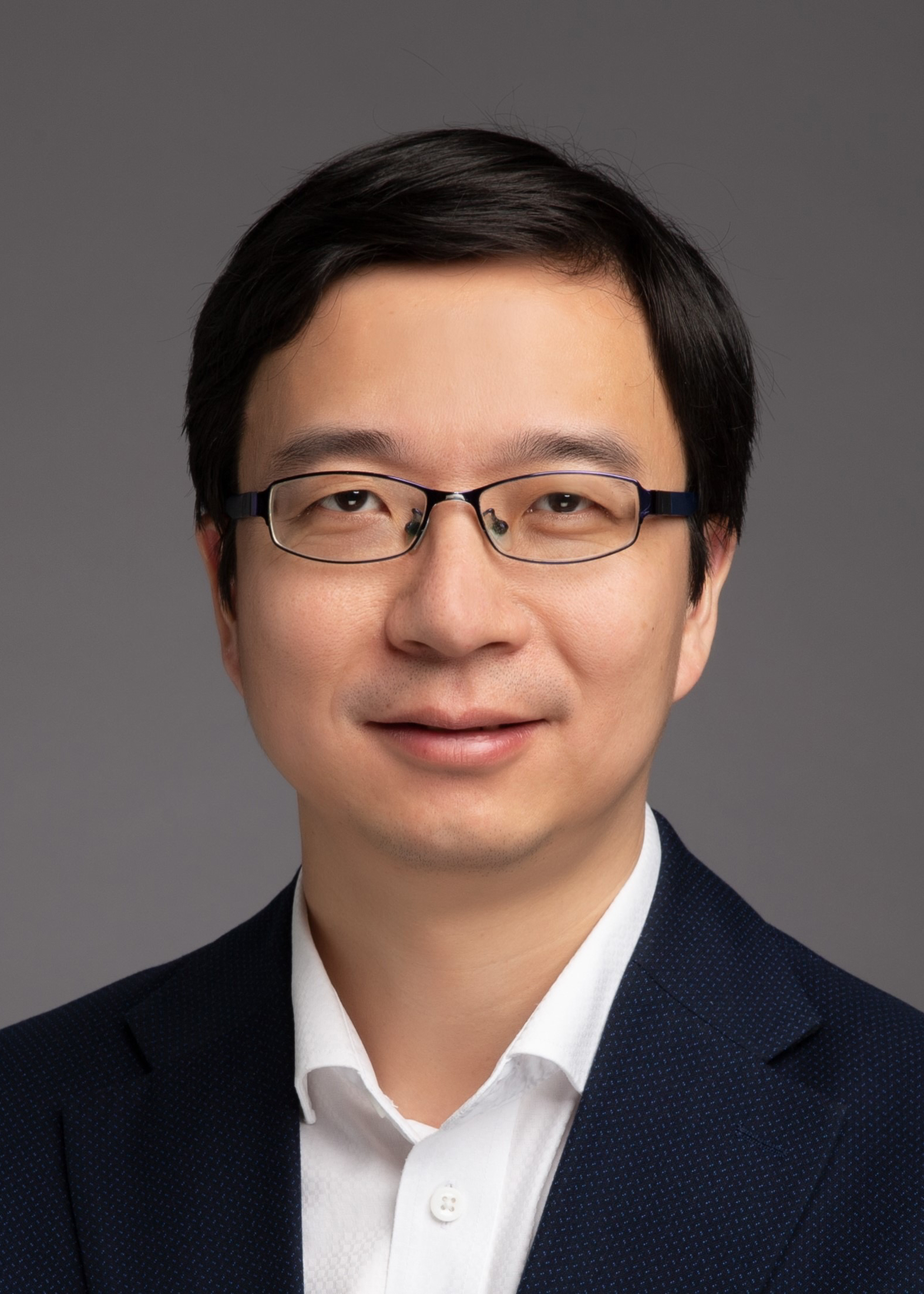}}]{Jiafeng Guo} received the PhD degree in computer software and theory from the University of Chinese Academy of Sciences. He is currently a Researcher of Institute of Computing Technology (ICT), Chinese Academy of Sciences (CAS) and a Professor of University of Chinese Academy of Sciences. He is the director of the CAS key lab of network data science and technology. His current research is focused on neural models for information retrieval (Neural IR) and natural language understanding.
\end{IEEEbiography}

\vskip -2\baselineskip plus -1fil

\begin{IEEEbiography}[{\includegraphics[width=1in,height=1.25in,clip,keepaspectratio]{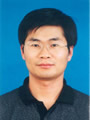}}]{Xueqi Cheng} (Senior Member, IEEE) is currently a Professor with the Institute of Computing Technology, Chinese Academy of Sciences. He has published more than 300 publications in prestigious journals and conferences. His research interests include network science, web search and data mining, big data processing, and distributed computing architecture.
\end{IEEEbiography}




\clearpage
\nobalance
\appendix

\begin{figure*}
    \centering
    \captionsetup{type=figure}
    
    \includegraphics[width=\linewidth]{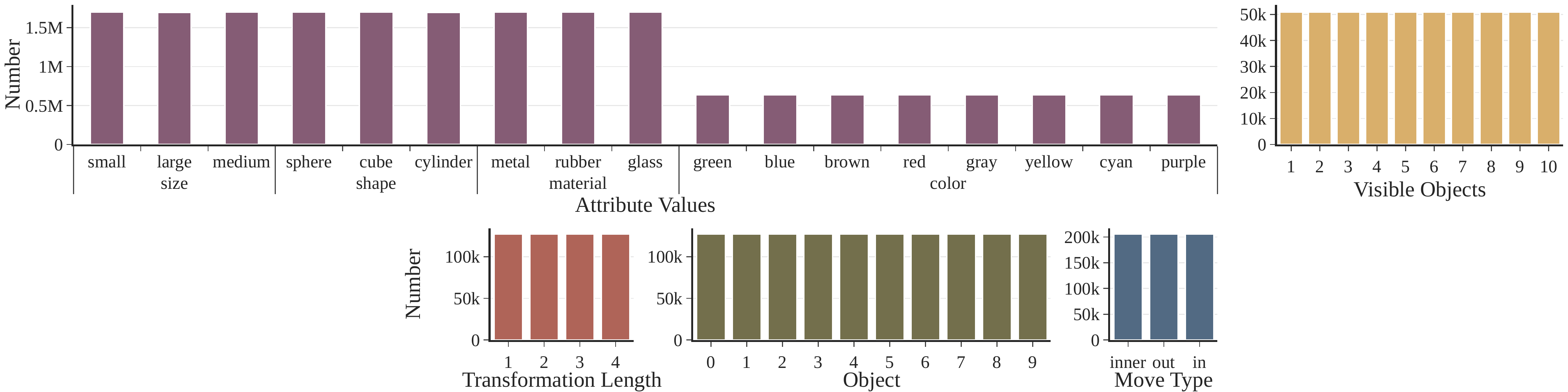}
    \captionof{figure}{The statistics of balanced factors in the TRANCE dataset. \textbf{Top Row:} attribute values and visible objects during sampling the initial scene graphs. \textbf{Bottom Row:} transformation length, object number, and move type during sampling transformation sequences.}
    \label{fig:statistics}
\end{figure*}%


\appendices

\section{Dataset Balance and Statistics on TRANCE} \label{sec:statistics}

Data balancing is an important factor that needs to be considered when constructing TRANCE. Major factors in TRANCE data, such as the length of a transformation sequence, that have the potential to be utilized by learners are balanced during the process of synthesizing the whole dataset. Without considering the rendering, the synthesizing process consists of two stages, i.e.~sampling an initial scene graph and sampling a transformation sequence to transform the initial scene graph into the final scene graph. In the following of this section, we first introduce the factors that are balanced in these two stages and then describe the method we used.

When sampling the initial scene graph, \textit{the number of visual objects} and \textit{the attribute values} of all objects are balanced strictly. Recall that the plane is separated into the visible area and invisible area and only objects in the visible area appear in the image of the initial state. The two diagrams on the top row of \cref{fig:statistics} show the statistics of these two factors.

When sampling the transformation sequence, we balance four factors in total. The first factor is \textit{the length of transformation} so that the numbers of samples with different transformation lengths are equal. The statistical result of the transformation length can be found on the left of the second row in~\cref{fig:statistics}. The other three factors refer to the elements of atomic transformations. For the part of the object, we balance the object number, and for the value, we balance the $n$-gram atomic transformation and the move type. \textit{The object number} is directly balanced over all samples and the result is shown in the middle of the second row in~\cref{fig:statistics}. As for the value, we consider the sub-sequence with the length $n$, which we called as \textit{$n$-gram atomic transformations}. The challenge of balancing this factor is that, for a specific initial scene graph, the availability of different atomic transformations is different. For example, changing the color of one object can always be successful, but changing the position of an object with a specific direction and step may be failed because of overlapping. Therefore, the concurrence of atomic transformations with low availability becomes rare. Intuitively, without balancing, four atomic transformations on position will be less possible than four atomic transformations on color existing in one sequence. Our balancing eliminates this potential bias and the statistics result is shown in \cref{tab:n_gram}. For each $n$-gram, the number of different options to be chosen is shown in the first row. For example, we have 33 different values so that the options of 1-gram are 33 and that of 2-gram is $33^2=1089$ and so on. We count the $n$-gram options with different sizes of sliding windows on all sampled transformation sequences. For example, we use a 2-length sliding window with 1 stride to count 2-gram atomic transformations on a 4-step transformation. Therefore, a 4-step contains three 2-gram atomic transformations. The remaining rows of \cref{tab:n_gram} are calculated on the counting results of options under each $n$-gram. From the table, the standard variance is very small compared to the mean value, which means the samples of different options under a specific $n$-gram is nearly equal. However, the size of TRANCE is 0.5 million, which is not enough to cover all 4-gram options, but the analysis of training data size has proved our data is sufficient for training a deep model. In conclusion, the $n$-gram atomic transformation is carefully balanced to eliminate the negative effect caused by the different availability of different atomic transformations. Additionally, we balance the \textit{move type} over all samples and the result is shown in the right of the second row in ~\cref{fig:statistics}.

The method we used to balance all the above factors is balanced sampling. The basic idea of this method is to change the sampling probability dynamically according to previously generated samples. \cref{alg:sample} shows how to sample an option from all available alternatives given the count table of previously generated all options.

\begin{table}[t]
\begin{center}

\caption{The statistics of the $n$-gram atomic transformations in TRANCE.}

\label{tab:n_gram}
\begin{small}
    \begin{tabular}{lrrrr}
    \toprule                
                   &   1-gram &  2-gram   & 3-gram &  4-gram \\
    \midrule
    options    & 33        & 1,089      & 35,937     & 1,185,921   \\
    \midrule
    min                 & 38,635     & 697       & 7         & 0         \\
    max                 & 38,638     & 708       & 15        & 3         \\
    median              & 38,636     & 703       & 11        & 0         \\
    mean                & 38,636     & 702.5     & 10.64     & 0.1075    \\
    std                 & 0.7714     & 2.2854    & 0.7880    & 0.3150    \\
    \bottomrule
    \end{tabular}
\end{small}
\end{center}
\end{table}





\begin{algorithm}[t]
\KwInput{available $k$ options $O = \{o_1, o_2, ... , o_k\}$, corresponding count table $N = \{n_1, n_2, ... , n_k\}$;}
\KwOutput{sampled option $o_r$;}
\KwParameter{tolerance $t = 0.1$ (default);}
\BlankLine

$n_{max}$ = max($n_1, n_2, ... , n_k$) \;
$c_i = n_{max} - n_i + t$ \;
$p_i = \frac{c_i}{\sum_{i=1}^{k}{c_i}}$ \;

$o_r$ = randomly sample an option from $\{o_1, o_2, ... , o_k\}$ with probability $\{p_1, p_2, ... , p_k\}$ \;
 \caption{Balanced Sampling}
 \label{alg:sample}
\end{algorithm}

\section{Implementation Details} \label{sec:detail}

The code for data generation is rewritten on the basis of the original code of CLEVR\footnote{\url{https://github.com/facebookresearch/clevr-dataset-gen}}.
As for training, we use PyTorch~\cite{NEURIPS2019_9015} as our deep learning framework for both TranceNets and TrancoNets. In the following, we introduce the implementation of our models and training process in detail. Code is publicly available at \url{https://github.com/hughplay/TVR}.

\subsection{Details of TranceNet Models}

\cref{tab:trance_model} shows the constitution of different TranceNet models. In the encoder part, both CNN$_{-}$ and CNN$_{\oplus}$ use a 4-layer CNN as the backbone of the encoder. The channel of four CNN layers is 16, 32, 32, 64, the kernel size is 5, 3, 3, 3, and all the strides are 2. The encoder backbone of ResNet$_{-}$, ResNet$_{\oplus}$, and DUDA is ResNet-18~\cite{he2016deep}, which we directly use the implementation given by PyTorch without pretrained parameters. As for BCNN, we use the VGG-18~\cite{simonyan2014very} implemented by PyTorch as the backbone of the encoder, which is consistent with the original paper~\cite{lin2015bilinear}. In the decoder part, the output of the encoder is first flattened and then encoded by a fully-connected layer to be a 128-dimension vector. This 128-dimension vector is then sent to a GRU network or a one layer transformer with the hidden size of 128. Finally, two 1-layer fully-connected layers are used to decode the object vector and the value of each step respectively. The dimension of the object vector is 19, 8 for the color, 3 for the size, 3 for the shape, 3 for the material, and 2 for the position. The dimension of the value output is 33.

\setlength{\tabcolsep}{5pt}
\begin{table}[t]
\begin{center}
\caption{The implementation of different TranceNet models on TRANCE.}
\label{tab:trance_model}
\begin{small}
    \begin{tabular}{lrrrr}
    \toprule                
     Model              &   Encoder           &  Decoder             & Parameters  \\
    \midrule
    CNN$_{-}$-G         & 4-layer CNN         & GRU           & 737K \\
    CNN$_{\oplus}$-G    & 4-layer CNN         & GRU           & 738K \\
    BCNN-G              & vgg11\_bn           & GRU           & 41M \\
    DUDA-G              & resnet18            & GRU           & 18M \\
    ResNet$_{-}$-G      & resnet18            & GRU           & 11M \\
    ResNet$_{\oplus}$-G & resnet18            & GRU           & 11M \\
    ResNet$_{-}$-T      & resnet18            & transformer           & 12M \\
    ResNet$_{\oplus}$-T & resnet18            & transformer           & 12M \\
    \bottomrule
    \end{tabular}
\end{small}
\end{center}
\end{table}
\setlength{\tabcolsep}{6pt}

The optimizer used for training is Adam~\cite{DBLP:journals/corr/KingmaB14}. The learning rate used by Adam is 0.001 in the beginning and is reduced to 0.0001 after 25 epochs. For the settings of Event and View, data is shared and the size of the training, validation and test set is 500,000, 2,000, and 8,000 respectively. For the setting of Basic, we collect all existing 1-step samples in data, and the size of the training, validation, and test set is 117,500, 2,000, and 8,000. All models are trained with 50 epochs on the training set and models that have the best results on the validation set are chosen to be evaluated on the test set to get the final results. In our experiments, images are resized to $120\times 160$ for fast training. Furthermore, by following the common practice on image augmentation~\cite{krizhevsky2012imagenet}, we apply a $0\sim5\%$ spatial translation to input image pairs during the training process.

\setlength{\tabcolsep}{5pt}
\begin{table}[t]
\begin{center}
\caption{The implementation of different TrancoNet models on TRANCO.}
\label{tab:tranco_model}
\begin{small}
    \begin{tabular}{lrrrrr}
    \toprule                
     \multirow{2}{*}{Model}              & \multirow{2}{*}{Encoder}           &  \multirow{2}{*}{Decoder}             &  \multicolumn{2}{c}{Parameters} \\
     \cmidrule{4-5}
     &&& Fixed & Trainable               \\
    \midrule

    RN101-G                 & RN101             & GRU           & 56.3M     & 6.3M \\ 
    ViT-B/32-T              & ViT-B/32          & Transformer   & 87.8M     & 7.4M \\
    ViT-B/16-T              & ViT-B/16          & Transformer   & 86.2M     & 7.4M \\ 
    RN101-T                 & RN101             & Transformer   & 56.3M     & 7.4M \\ 
    \bottomrule
    \end{tabular}
\end{small}
\end{center}
\end{table}
\setlength{\tabcolsep}{6pt}

\subsection{Details of TrancoNet Models}

\cref{tab:tranco_model} shows the constitution of different TrancoNet models. In the encoder part, we test three image encoders from the released CLIP models \footnote{\url{https://github.com/openai/CLIP}}, including RN101, ViT-B/32, and ViT-B/16. The initial and the final states are separately encoded by CLIP image encoder, resulting in two 512-dimension vectors. These two vectors are then sent into a 1-layer transformer to obtain $\vb*{h}$, to be the output of the encoder. In the decoder part, in each step, the previous steps of atomic transformations, which are video clips, are first sent into CLIP image encoder to obtain clip vectors, and then encoded by a 1-layer transformer with $\vb*{h}$ in the way of cross attention. RN101-G uses GRU as the decoder, and the difference is only the last step of atomic transformation is sent to encoding. Finally, the output vector of this step is used to find the most similar video clip from the candidate. Each video clip from the candidate is sent into the image encoder of CLIP like the way for previous steps, in which three random frames are selected to encode separately and then averaged, as introduced in~\cite{PortilloQuintero2021ASF}.

Mixed precision is applied during training and the time for training a model ranges from 8 to 20 GPU hours, according to different backbones and pre-training settings. AdamW~\cite{Loshchilov2019DecoupledWD} is used as an optimizer, and the learning rate gradually warmup~\cite{vaswani2017attention} to $2^e{-4}$ in the first 2,000 steps. $\tau$ is simply set to 1 for all experiments. Techniques are used to avoid model over-fitting such as sparse temporal sampling~\cite{wang2016_TemporalSegmentNetworks}, data augmentation such as random cropping and random flipping, dropout, etc.

\section{Analysis of the Data Size on TRANCE}

\begin{figure}[t]
    \centering
    \includegraphics[width=0.83\linewidth]{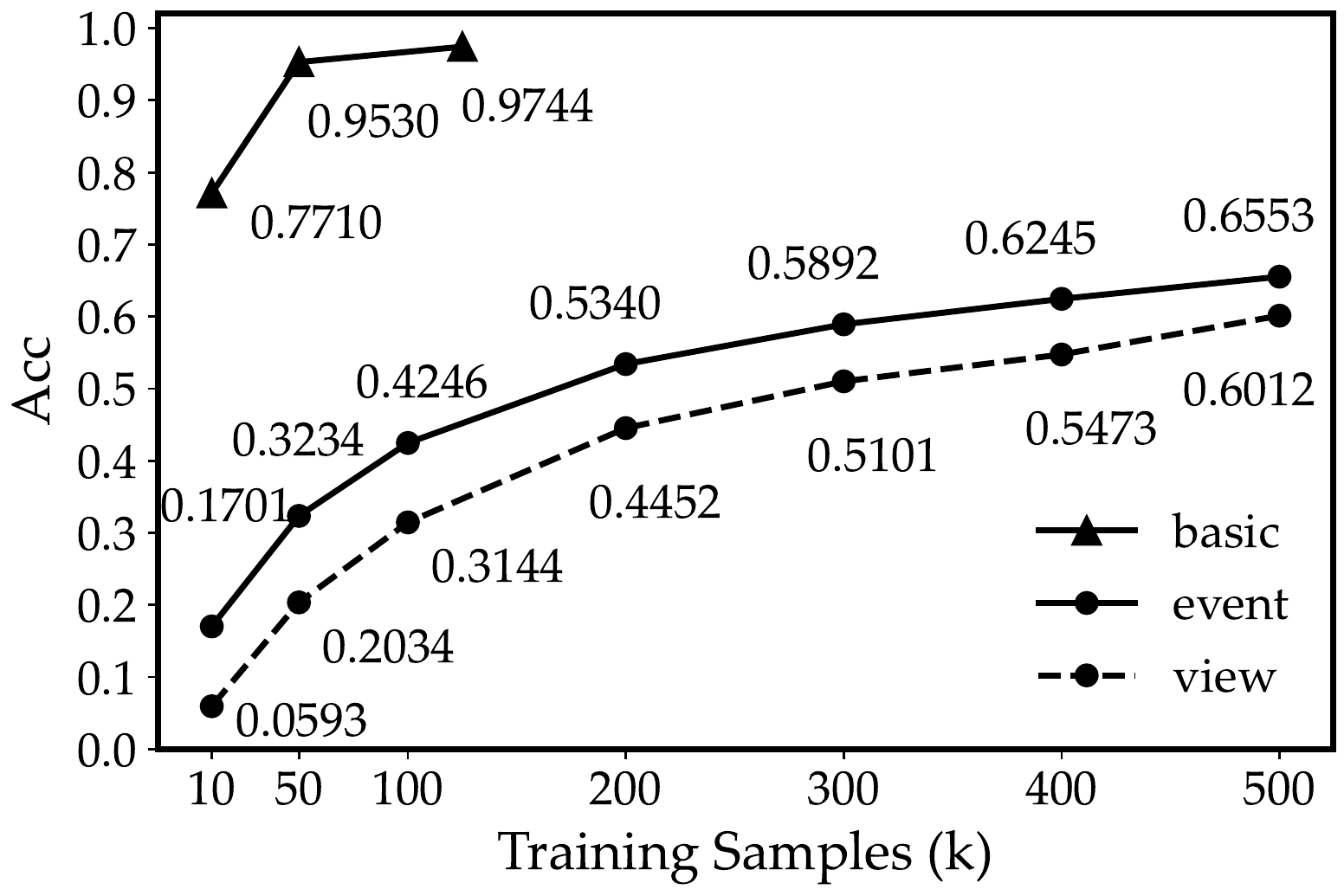}
    \caption{Results of ResNet$_{-}$-T with different training size.}
    \label{fig:size}
\end{figure}

It is a common question whether the training data size is large enough for training a deep model. We study the influence of this factor on ResNet$_-$-T. From~\cref{fig:size}, we can see that more training samples bring significant benefits when the number is less than 50k on Basic and 200k on Event and View. After that, the benefits become smaller and smaller. Those results are consistent with the common knowledge that relatively large data is required to well train a deep model. These results also show the data size of TRANCE is large enough for deep models.

\begin{figure}[t]
    \centering
    \includegraphics[width=0.95\linewidth]{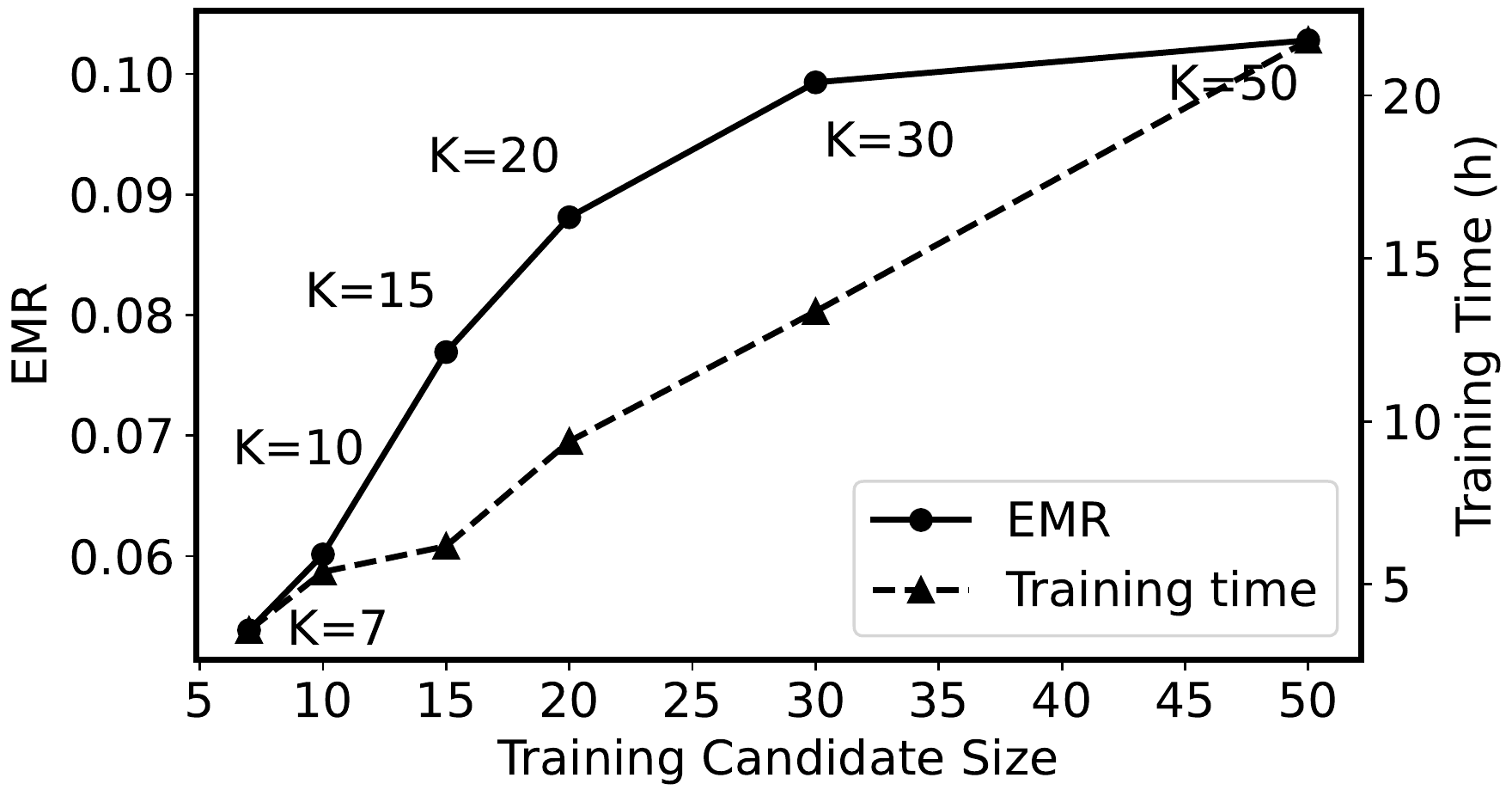}
    \caption{The trade-off between the candidate size and performance.}
    \label{fig:candidates}
\end{figure}

\section{Analysis of the Candidate Size on TRANCO}
\label{sec:candidate_size}

The selection of training candidate $K$ is a balance between the performance and computation complexity. We change $K$ and train multiple ViT-B/32-T models. From \cref{fig:candidates}, results show larger candidate size brings significant benefits when the size is smaller than 20, but the benefit gradually becomes smaller and smaller. However, the computational cost increases linearly according to the candidate size and finally leads to a prohibitive computation cost. Therefore, after considering the trade-off between performance and computational costs, our major experiments choose 20 as an efficient choice.

\section{Human Test System for TRANCE} \label{sec:human_test}

\begin{figure*}[t]
    \centering
    \includegraphics[width=0.75\linewidth]{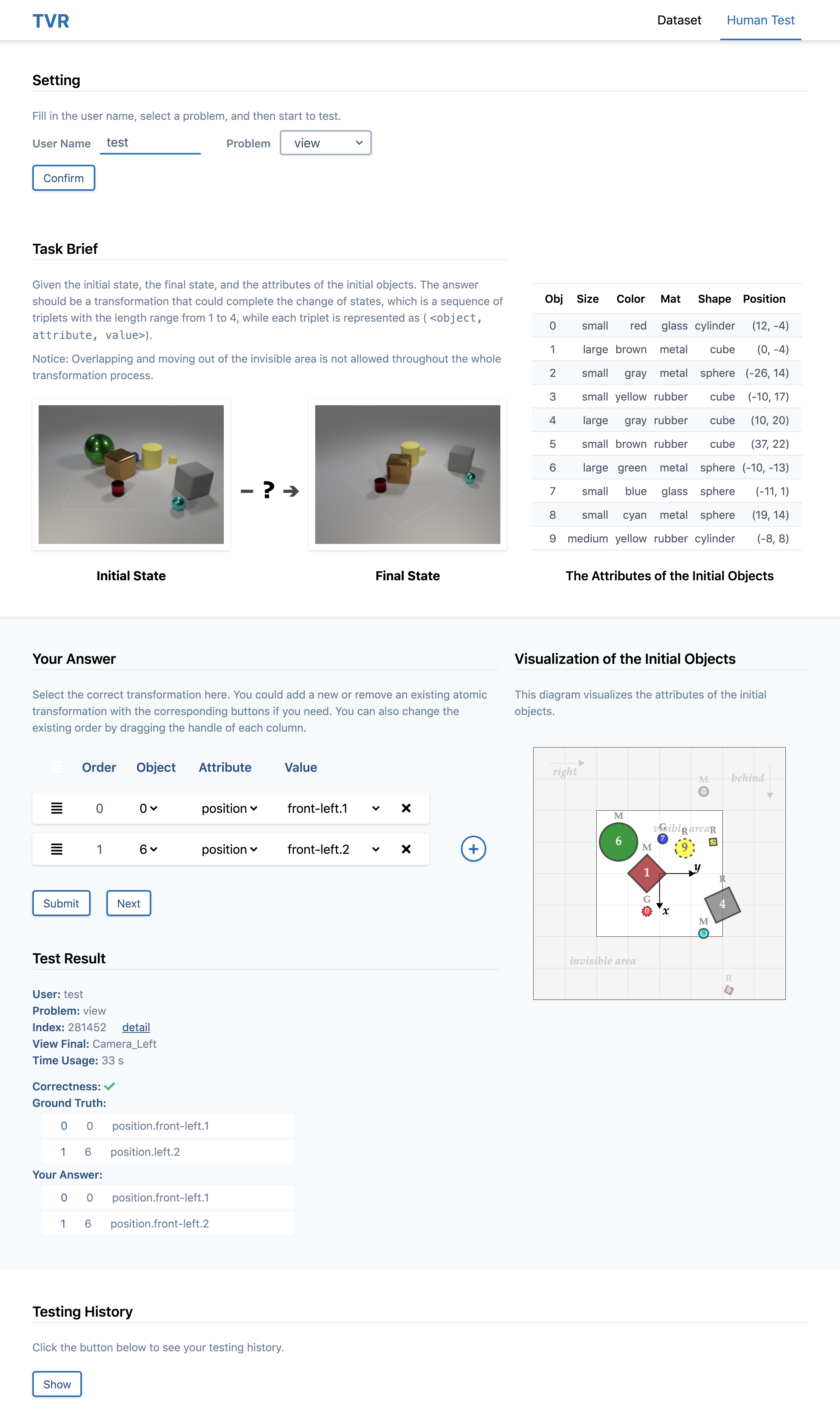}
    \caption{Human test system.}
    \label{fig:human_test}
\end{figure*}

To collect results from humans, we build a web-based test system. \cref{fig:human_test} shows the GUI of this system. The whole testing process is described in the following steps. First, a human tester is told to be familiar with the system by trying a few examples with guidance. After that, the tester changes the user name and the target problem to start testing. During the testing of each sample, the tester should select the correct atomic transformations arranged in a feasible order after observing the initial, the final state, and the attributes of the initial objects. To reduce the time usage, we also provide the visualization of the initial objects for testers. After that, the tester can submit the answer and start to answer the next sample. After completing all test samples, the tester can see his or her test result by clicking the button under the testing history. The code for the human test system is also publicly available.

\begin{figure}[t]
    \centering
    \centerline{\includegraphics[width=\linewidth]{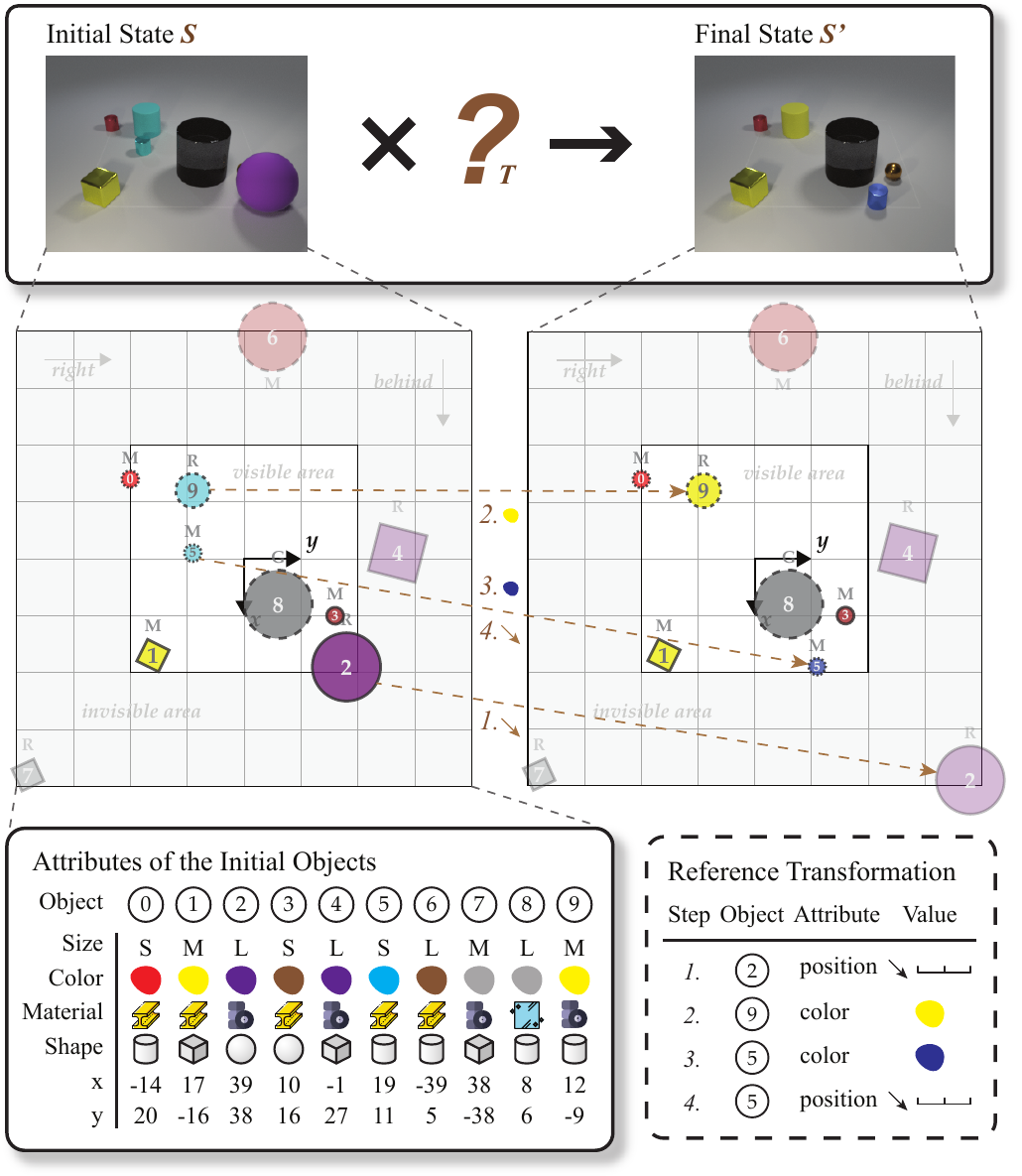}}
    \caption{An example from the Event setting.}
    \label{fig:example}
\end{figure}

\section{Examples of TRANCE} \label{sec:examples}

We show an example from the Event setting in \cref{fig:example} to better explain how TRANCE is like. In this example, there are seven objects in the visible area in the initial state and six objects in the visible area in the final state. If you look closely, you will find there are four changes, which are shown as the reference transformation. One thing that should be noted is that the order between two position transformations is not reversible: moving object \#5 before moving object \#2 leads to object overlapping which is not allowable.

The remaining pages show extra examples from the three settings of TRANCE, i.e.~Basic, Event, and View. In each sample, the initial state, the final state, and the attributes of the initial objects are given. In the View setting, the view angle of the final state is only randomly selected from the Left and Right, since samples with the Center view are similar to the samples from the Event setting. Besides, for each sample, an additional diagram is provided to visualize the attributes of the initial objects. At last, we show the reference transformation.

When moving an object from the visible area into the invisible area, any directions and steps that could cause the same effect without making objects overlap are accepted. This is implemented by only comparing the visible objects' attribute values of the final states in the evaluation system.

\begin{figure*}[ht]
    \centering
    \includegraphics[width=0.95\linewidth]{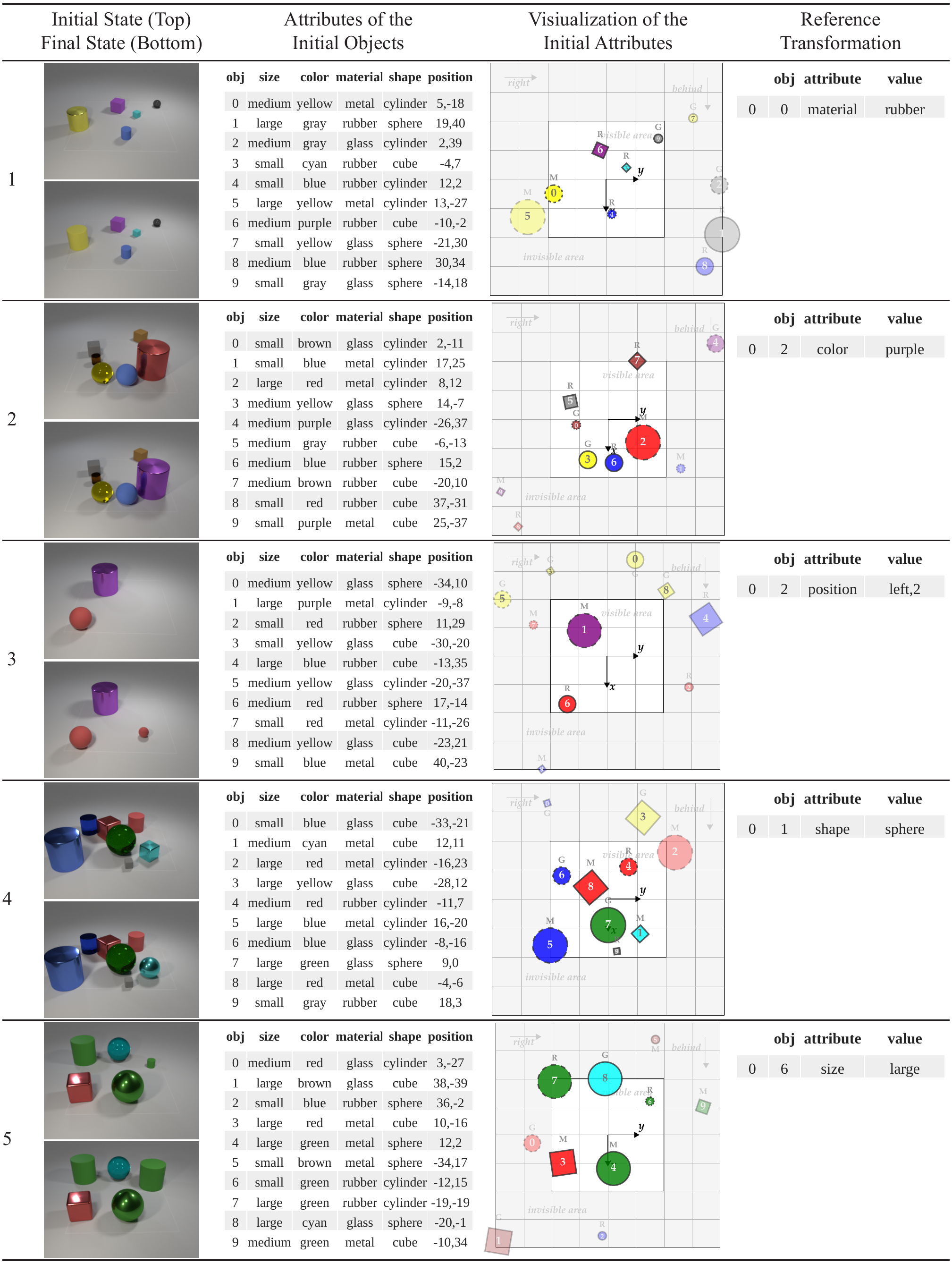}
    \caption{Examples from the Basic setting.}
    \label{fig:example_basic}
\end{figure*}

\begin{figure*}[ht]
    \centering
    \includegraphics[width=0.95\linewidth]{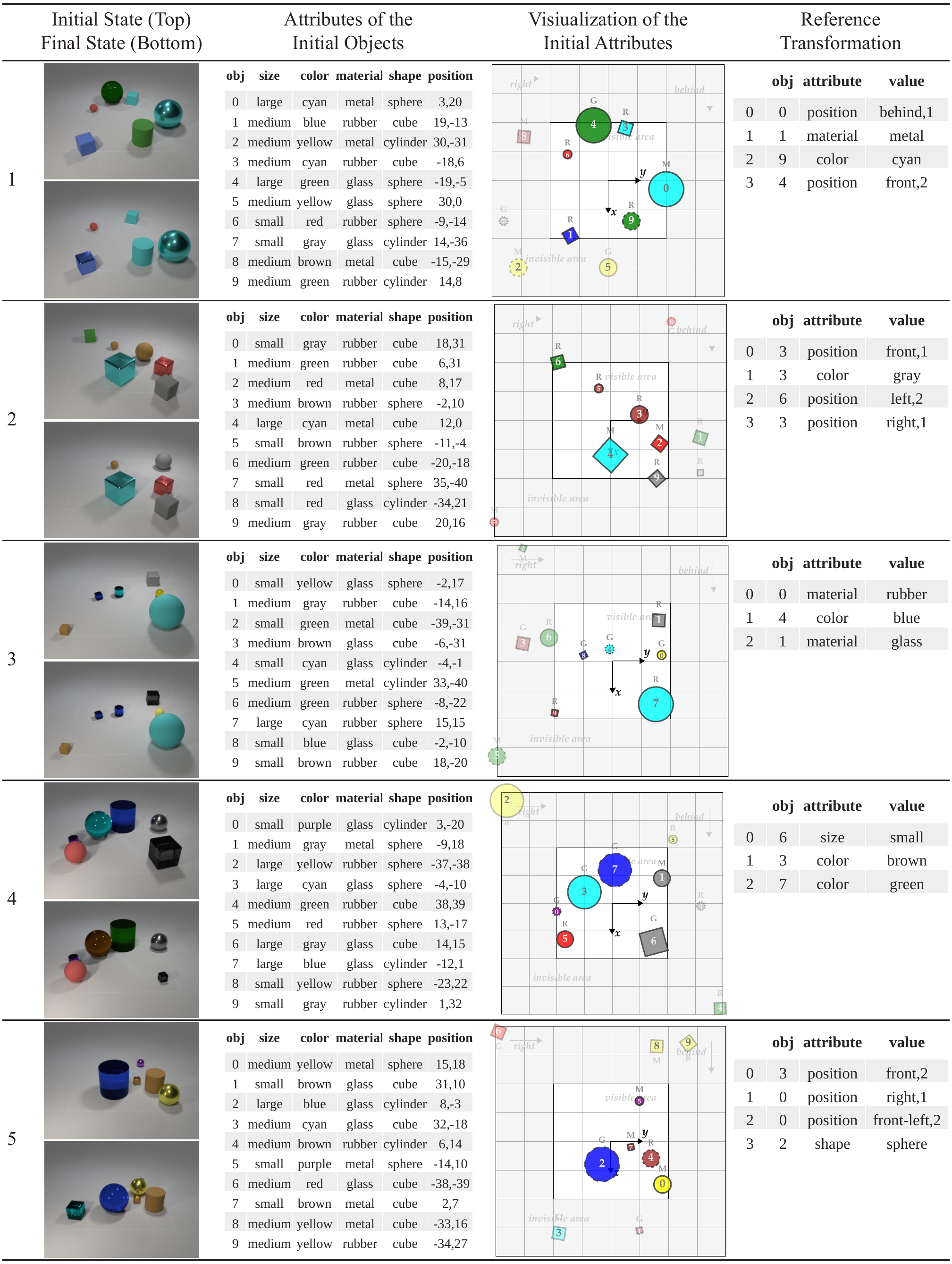}
    \caption{Examples from the Event setting.}
    \label{fig:example_event}
\end{figure*}

\begin{figure*}[ht]
    \centering
    \includegraphics[width=0.95\linewidth]{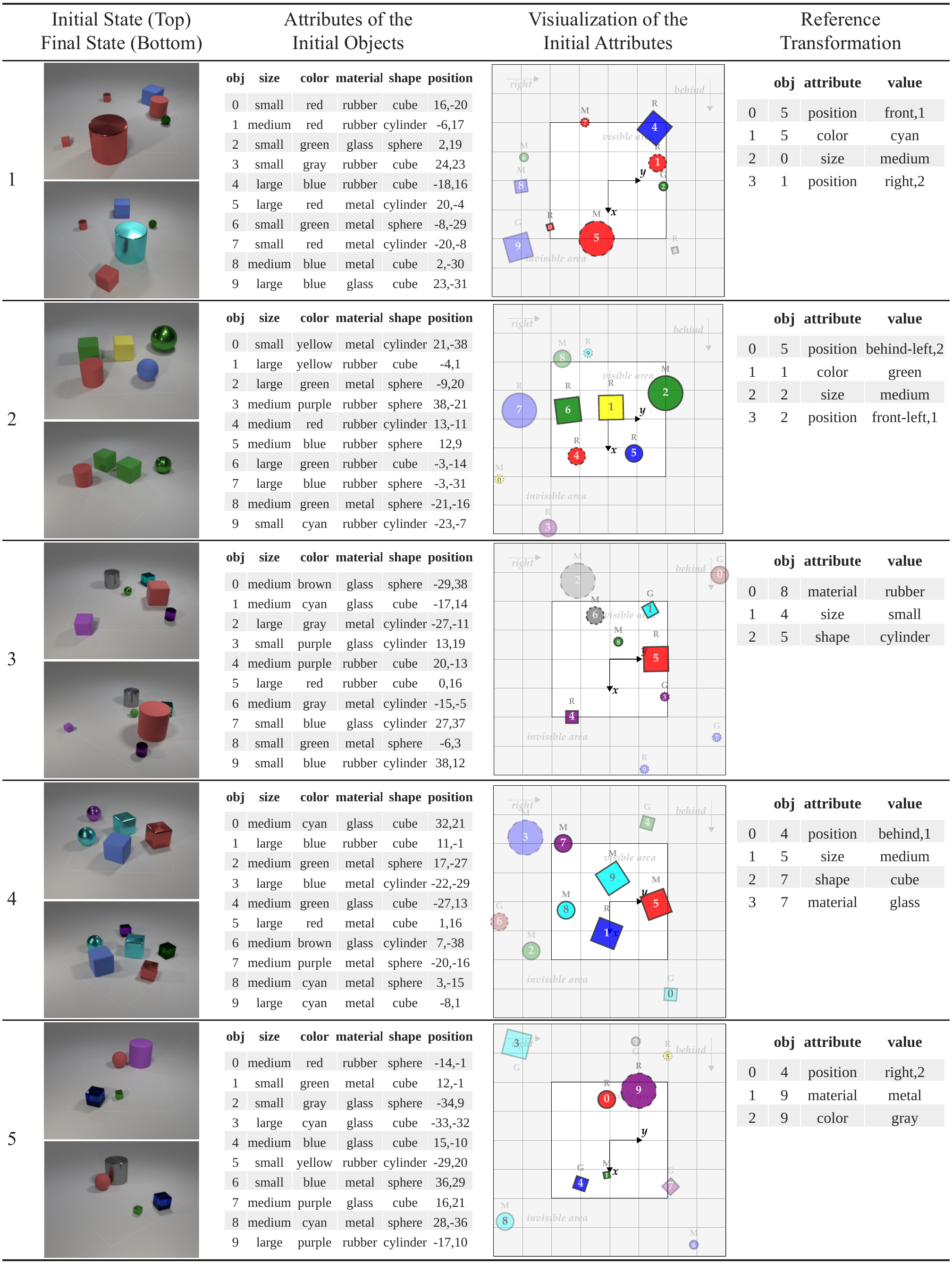}
    \caption{Examples from the View setting.}
    \label{fig:example_view}
\end{figure*}

\clearpage


\end{document}